\documentclass[10pt,twocolumn,letterpaper]{article}

\usepackage{cvpr}              

\usepackage[dvipsnames]{xcolor}

\usepackage{lineno}
\newcommand{\expnum}[2]{{#1}\mathrm{e}{-#2}}

\definecolor{cvprblue}{rgb}{0.21,0.49,0.74}
\usepackage[pagebackref,breaklinks,colorlinks,allcolors=cvprblue]{hyperref}
\usepackage{graphicx}
\usepackage{amsmath}
\usepackage{amssymb}
\usepackage{booktabs}
\usepackage{ragged2e}

\usepackage{color}
\usepackage{amssymb}

\usepackage{lipsum}
\usepackage{xcolor}
\usepackage{tabularx}
\usepackage{multirow}
\usepackage{enumitem}
\usepackage{bbm}
\usepackage{wrapfig}
\usepackage{setspace}
\usepackage{footnote}

\usepackage{colortbl} 
\usepackage{pifont}  
\usepackage{mathtools}  
\usepackage[font=small]{caption}  
\usepackage{blindtext}  
\usepackage{stmaryrd} 
\usepackage[accsupp]{axessibility}  
\usepackage{mathtools}
\usepackage{subfiles}
\usepackage{afterpage}

\usepackage{algorithm}
\usepackage{algpseudocode}
\usepackage{adjustbox}

\usepackage{anyfontsize}
\usepackage{afterpage}
\usepackage{xcolor,pifont}

\definecolor{amethyst}{rgb}{0.6, 0.4, 0.8}

\definecolor{grey}{rgb}{0.9, 0.9, 0.9}
\newcommand{\ccol}{\cellcolor{grey}}
\newcommand*\colourcheck[1]{%
  \expandafter\newcommand\csname #1check\endcsname{\textcolor{#1}{\ding{51}}}%
}

\colourcheck{blue}
\colourcheck{green}
\colourcheck{red}
\newcommand{\cmark}{\ding{51}}%
\newcommand{\xmark}{\ding{55}}%

\def\ie{\emph{i.e.}}
\def\eg{\emph{e.g.}}
\def\etal{\emph{et al.}}

\definecolor{bright_red}{rgb}{0.97, 0.9, 0.9}
\definecolor{blk}{rgb}{0, 0, 0}
\definecolor{grn}{rgb}{0, 0.6, 0}
\definecolor{mgt}{rgb}{0.8, 0.1, 0.8}
\definecolor{darkblue}{rgb}{0.2, 0.2, 0.8}
\definecolor{lblue}{rgb}{0.2, 0.2, 1.0}
\definecolor{orange}{rgb}{1.0, 0.5, 0.2}
\definecolor{goldenrod}{rgb}{0.85, 0.65, 0.33}
\definecolor{darkred}{rgb}{0.75, 0.0, 0.0}

\newcommand{\bs}[1]{{\color{black}{#1}}}

\makeatletter
\newcommand{\multiline}[1]{%
  \begin{tabularx}{\dimexpr\linewidth-\ALG@thistlm}[t]{@{}X@{}}
    #1111
  \end{tabularx}
}
\makeatother

\def\ie{\emph{i.e.}}
\def\eg{\emph{e.g.}}

\definecolor{brown}{rgb}{0.85, 0.15, 0.15}
\definecolor{purp}{rgb}{0.95, 0.16, 0.65}
\definecolor{purpc}{rgb}{0.95, 0.36, 0.65}
\definecolor{orange}{rgb}{1.0, 0.5, 0.0}
\definecolor{blue}{rgb}{0.0, 0.5, 1.0}
\definecolor{green}{rgb}{0, 0.8, 0}
\definecolor{lgreen}{rgb}{0.6, 0.8, 0}
\definecolor{red}{rgb}{0.8, 0, 0}
\definecolor{darkblue}{rgb}{0.2, 0.2, 0.8}
\definecolor{brinkpink}{rgb}{0.98, 0.38, 0.5}
\definecolor{cadmiumred}{rgb}{0.89, 0.0, 0.13}
\definecolor{ceruleanblue}{rgb}{0.16, 0.32, 0.75}
\definecolor{dandelion}{rgb}{0.94, 0.88, 0.19}
\definecolor{bostonuniversityred}{rgb}{0.8, 0.0, 0.0}
\definecolor{brown(web)}{rgb}{0.65, 0.16, 0.16}
\definecolor{cornellred}{rgb}{0.7, 0.11, 0.11}

\newcolumntype{C}[1]{>{\centering\let\newline\\\arraybackslash\hspace{0pt}}p{#1}}

\def\etal{\emph{et al.}}

\definecolor{grey}{rgb}{0.9, 0.9, 0.9}

\newcommand{\hyperfootnote}[1][]{\def\ArgI\hyperfootnoteRelay}
\newcommand\hyperfootnoteRelay[2][]{\href{#1#2}{\ArgI}\footnote{\href{#1#2}{#2}}}

\def\paperID{5016} 
\def\confName{CVPR}
\def\confYear{2025}

\title{Learning Audio-guided Video Representation with Gated Attention for Video-Text Retrieval}

\author{Boseung Jeong$^{1}$
\hspace{9mm}
Jicheol Park$^{2}$
\hspace{9mm}
Sungyeon Kim$^{1}$
\hspace{9mm}
Suha Kwak$^{1,2}$\vspace{2mm}\\
$^1$Dept. of CSE, POSTECH \qquad \ \       
$^2$Graduate School of AI, POSTECH \\
{\tt \{\small{boseung01, jicheol, sungyeon.kim, suha.kwak}\}@postech.ac.kr}\\
{\tt \small \url{http://cvlab.postech.ac.kr/research/AVIGATE}}
}

\begin{document}

\maketitle

\begin{abstract}
Video-text retrieval, the task of retrieving videos based on a textual query or vice versa, is of paramount importance for video understanding and multimodal information retrieval. 
Recent methods in this area rely primarily on visual and textual features and often ignore audio, although it helps enhance overall comprehension of video content.
Moreover, traditional models that incorporate audio blindly utilize the audio input regardless of whether it is useful or not, resulting in suboptimal video representation. 
To address these limitations, we propose a novel video-text retrieval framework, Audio-guided VIdeo representation learning with GATEd attention (AVIGATE), that effectively leverages audio cues through a gated attention mechanism that selectively filters out uninformative audio signals.
In addition, we propose an adaptive margin-based contrastive loss to deal with the inherently unclear positive-negative relationship between video and text, which facilitates learning better video-text alignment.
Our extensive experiments demonstrate that AVIGATE achieves state-of-the-art performance on all the public benchmarks.
\end{abstract}

\section{Introduction}
\label{sec:intro}

Video-text retrieval, the task of finding videos relevant to text query or vice versa, has gained significant interest due to its diverse applications in video understanding and multi-modal information retrieval. 
\begin{figure}[t!]
    \centering
    \includegraphics[width=0.99\linewidth]{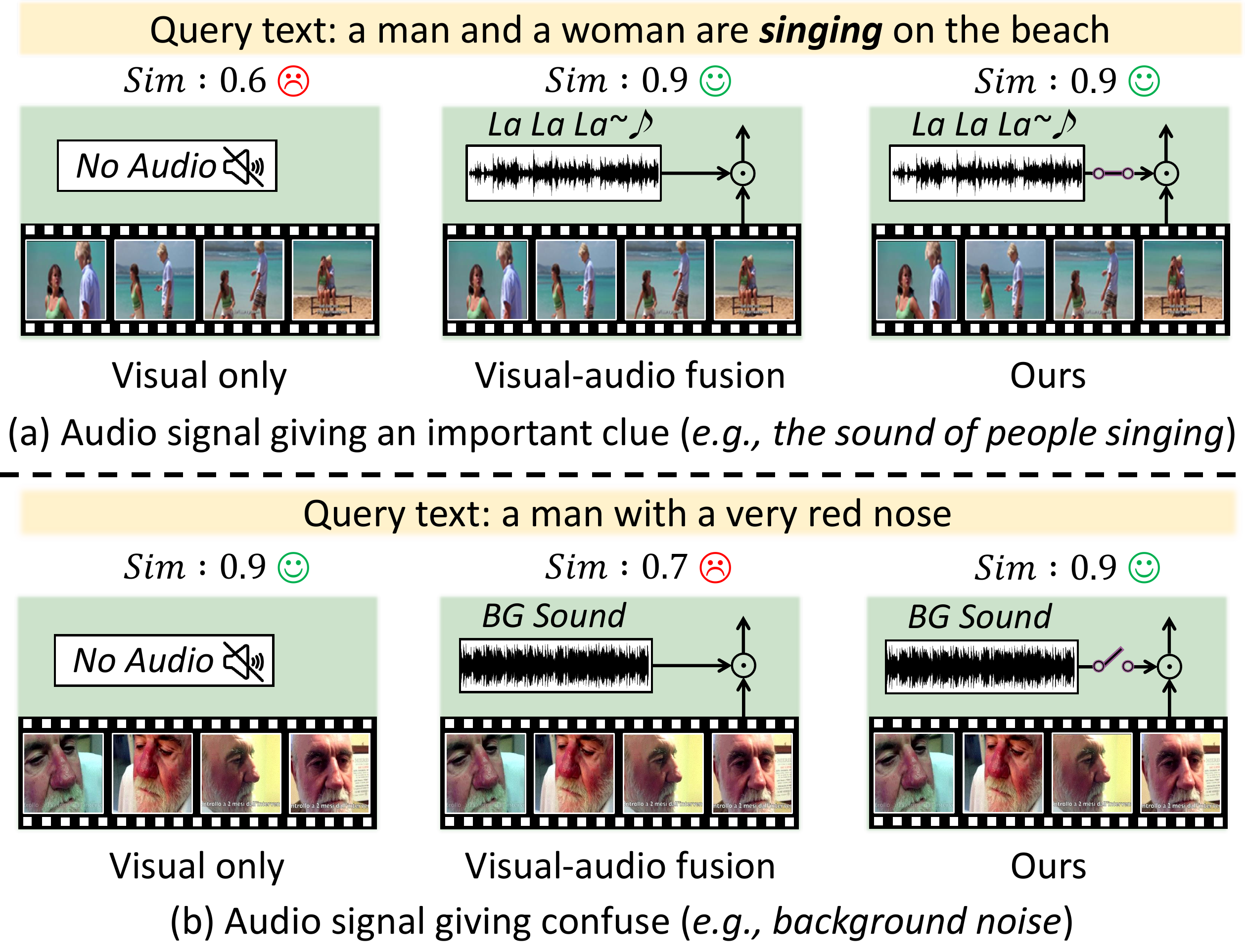}
    \vspace{-2mm}
    \caption{Comparative illustration of different scenarios using visual-only, audio-video fusion, and our proposed gated fusion approach. (a) In cases where the audio signal provides valuable information, the audio-video fusion and our gated fusion achieve high similarity scores. (b) When the audio signal is misleading, traditional fusion methods degrade performance. In contrast, our gated fusion mechanism successfully filters irrelevant audio cues, maintaining a high similarity score like the visual-only case.}
    \label{fig: teaser}
    \vspace{-3mm}
\end{figure}
Most existing methods in this field have primarily focused on leveraging visual information from videos and textual information from accompanying metadata or captions~\cite{clip4clip,clip2video,camoe,xclip,xpool,ts2,uatvr,prost,ucofia}. 
While these methods have greatly advanced video-text retrieval, they still have a significant limitation: they neglect audio information of video that conveys `\textit{invisible but audible}' cues providing essential context such as speaker identity, background noises, or emotional nuances, which are crucial for a comprehensive understanding of video.
Leveraging the audio information in video representation learning has potential to substantially improve retrieval performance by offering a richer multi-modal representation of video, as shown in Figure~\ref{fig: teaser}(a).

Nevertheless, only a few recent studies~\cite{eclipse,tefal} have explored the use of audio information in video-text retrieval tasks. ECLIPSE~\cite{eclipse} employs a cross-modal attention mechanism to fuse audio and visual modalities, creating a unified representation. TEFAL~\cite{tefal}, on the other hand, proposes text-conditioned feature alignment based on the cross-attention mechanism for audio and textual modalities as well as for visual and textual modalities.
These methods commonly assume that audio contributes positively to enhancing the video representation, but this is not always the case.
When irrelevant audio is processed jointly with visual information~(\eg, irrelevant background music and noise), it could impair the video representation and negatively affect cross-modal alignment, as shown in Figure~\ref{fig: teaser}(b). 
ECLIPSE and TEFAL do not address this issue and blindly exploit the audio modality.
Moreover, TEFAL introduces substantial computational demands on retrieval systems since both the video (also audio) and text description must be processed together to generate the representation of both video frame and audio, necessitating reprocessing of the entire database each time a new query is received.

To address these issues, we introduce a novel framework, \textbf{A}udio-guided \textbf{VI}deo representation learning with \textbf{GATE}d attention~(\textbf{AVIGATE}), that dynamically determines whether audio input is useful and enables efficient retrieval by processing video (including audio) and text description independently; its overall architecture is illustrated in Figure~\ref{fig: arch}.
AVIGATE is built on three encoders for each modality: Audio Spectrogram Transformer (AST)~\cite{ast} for the audio input and two CLIP encoders~\cite{clip}, one for visual input and the other for textual input.
The model begins by encoding audio using AST to produce dense audio embeddings. 
These embeddings are then processed by an audio resampler to preserve rich audio information while reducing redundancy by resampling to a fixed number of embeddings. 
Video is represented by frame embeddings given by the CLIP image encoder, while the text is encoded by the CLIP text encoder. 

Audio and frame embeddings are then fused by a gated fusion transformer. 
This module integrates audio and frame embeddings by employing an adaptive gating function that regulates the influence of each modality. 
To be specific, the gate function dynamically determines the contribution of audio embeddings, allowing the transformer to leverage complementary audio information when relevant while maintaining the frame embeddings by minimizing the effect of potentially noisy audio.
The output of the gated fusion transformer, \ie, the final video representation, is then aligned with the text embedding.

To improve the discriminative capability of video-text alignment, we propose an adaptive margin-based contrastive learning approach that extends the traditional contrastive loss~\cite{Sohn_nips2016,oord2018representation} by incorporating an additional margin for each negative pair.
The margin is determined based on the intra-modal dissimilarity within both the textual and visual modalities, allowing for implicit exploration of inherent relationships among semantically similar pairs.
The input similarity score between the video representation and the text embedding used in the proposed loss is computed through a multi-grained alignment scheme~\cite{xclip,prost,ucofia,uatvr} that enables capturing both the overall semantic context and fine details.
Consequently, the proposed loss with the multi-grained alignment scheme encourages learning a more discriminative and generalizable cross-modal embedding space, leading to a better video-text alignment. 

Our method was evaluated and compared with prior work on two public benchmarks~\cite{msrvtt,vatex} where it outperformed all existing methods while enabling an efficient retrieval.
The main contribution of this paper is three-fold:
\begin{itemize}[leftmargin=*, topsep=1mm]
\item We propose AVIGATE, an effective audio-video fusion framework that fuses audio and visual information while dynamically determining whether the audio is valuable. 
\item We propose a new adaptive margin-based contrastive loss, considering inherent relationships between semantically similar pairs so that the cross-modal embedding space becomes more discriminative and better generalizes. 
\item  Our method achieves the best performance on \bs{three} public benchmarks for video-text retrieval and ensures a high retrieval efficiency during testing.
\end{itemize}

\section{Related Work}
\subsection{Video-Text Retrieval}
Video-text retrieval is a fundamental topic in the vision-language domain aimed at retrieving the most semantically relevant video for a given text query.
Early work in this field primarily employed dense fusion mechanisms for cross-modal alignment, aiming to tightly integrate features across modalities~\cite{ctsan, jsf}.
However, with the introduction of large-scale text-video datasets, more recent approaches have shifted toward end-to-end pre-training strategies that allow models to learn video and text features jointly.

To further these advancements, notable methods such as ClipBERT~\cite{clipbert} and Frozen~\cite{frozen} have proposed efficient training techniques like sparse sampling and curriculum learning, enhancing the feasibility of handling large datasets.
Additionally, recent methods have leveraged models pre-trained on massive image-text pairs, such as CLIP~\cite{clip}, to benefit from their robust visual-textual alignment capabilities.
CLIP4Clip~\cite{clip4clip} applies frame-level alignment within CLIP’s pre-trained feature space to significantly improve video retrieval performance.

Expanding on CLIP4Clip, recent studies emphasize achieving precise alignment between visual and textual modalities.
X-Pool~\cite{xpool} calculates cross-attention weights between text and video frames to extract text-conditioned video representations, where the video encoding process must involve both modal data, introducing significant computational costs.
Meanwhile, a series of studies~\cite{xclip,ucofia,uatvr,prost} have explored and exploited a multi-grained alignment scheme to achieve more accurate retrieval.  
UCOFiA~\cite{ucofia} employs a hierarchical alignment strategy across patch-word, frame-sentence, and video-sentence levels.
UATVR~\cite{uatvr} that begins with token-wise word-frame matching addresses semantic uncertainty in video-text pairs by introducing a distribution matching method that models modality-specific distributions, enabling more robust and adaptive matching.
Following this spirit, our method, built on the principles of CLIP4Clip, also exploits a multi-grained alignment scheme in a na\"ive way. 
\subsection{Audio-Enhanced Video-Text Retrieval}
Although video-text retrieval has advanced significantly, there remains room for improvement by incorporating audio, readily available yet often overlooked in video content.
An early study by Liu \etal,~\cite{collaboexperts} uses audio information for video-text retrieval by integrating the pre-trained representation of audio experts with that of other modalities.

More recent studies have focused on learning representations that fuse audio with other modalities.
ECLIPSE~\cite{eclipse} introduces a fusion method using cross-attention between audio and video to generate audio-guided video representations.
On the other hand, TEFAL~\cite{tefal} applies cross-attention between text and audio, as well as between text and video, to produce text-guided audio and video representations.
This process requires a large burden of computational costs, hindering efficient retrieval.

Furthermore, the prior fusion methods have limitations, as they fail to handle irrelevant audio, like background noises, which can degrade video representations and hinder cross-modal alignment.
To tackle these issues, we present a new fusion method that dynamically assesses the relevance of audio input and enables an efficient retrieval system by independently processing video and text descriptions.
\label{sec:relate}


\section{Proposed Method}
\label{sec:method}

\begin{figure*}[t!]
    \centering
    \vspace{-1mm}
    \includegraphics[width=0.99\linewidth]{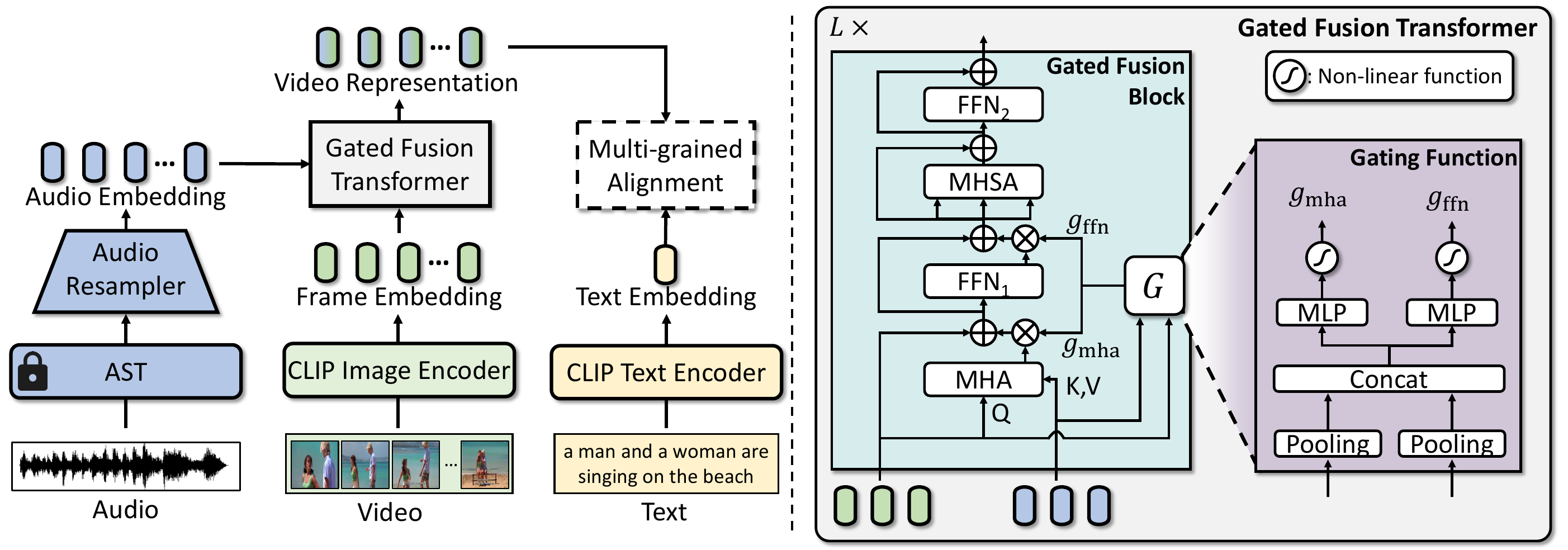}
    \vspace{-2mm}
    \caption{($Left$) The overall architecture of AVIGATE. Audio input is processed through an Audio Spectrogram Transformer (AST) and further refined by an audio resampler to generate fixed-size audio embeddings. Frame embeddings are derived from the video using a CLIP Image Encoder, while the text embedding is extracted by the CLIP Text Encoder. These audio and frame embeddings are fused by a gated fusion transformer, which dynamically determines the contribution of audio. 
    The final video representation is aligned with the text embedding using a multi-grained alignment scheme, facilitating an effective video-text retrieval process.
    ($Right$) The gated fusion transformer consists of a gated fusion block and a gating function.
    }
    \label{fig: arch}
    \vspace{-3.5mm}
\end{figure*}
This section presents details of our video-text retrieval framework, dubbed AVIGATE; its overall architecture is depicted in Figure~\ref{fig: arch}. 
We first describe the embedding extraction process for each modality in ~\cref{sec:extraction}, and then elaborate on the gated fusion transformer in~\cref{sec:gate}. 
Finally, the adaptive margin-based contrastive learning strategy is introduced in~\cref{sec:margin_loss}.
\subsection{Embedding Extraction}
\label{sec:extraction}

AVIGATE utilizes three pre-trained encoders to process each modality: Two CLIP encoders~\cite{clip} for the video frames and the text descriptions, respectively, and Audio Spectrogram Transformer (AST)~\cite{ast} for the audio signals.

\noindent\textbf{Frame Embeddings:} We utilize the pre-trained CLIP image encoder~\cite{clip} to extract the frame embeddings of each video.
Given an input video $V$, $N$ frames are first uniformly sampled denoted as $V=[F^1, F^2, \dots, F^N]$. 
Each frame is divided into non-overlapping patches, which are then transformed using a linear projection. 
A [\texttt{CLS}] token is prepended to this sequence of patches, and the combined sequence is fed to the CLIP image encoder. 
The output corresponding to the [\texttt{CLS}] token is taken as the frame embedding.
Finally, these frame embeddings are concatenated to form a sequence of frame embeddings, $\mathbf{f}=[\mathbf{f}^1, \mathbf{f}^2, \dots, \mathbf{f}^N]\in\mathbb{R}^{N\times D}$, where $D$ denotes the dimension of the embeddings. 

\noindent\textbf{Text Embeddings:} The text embedding extraction is carried out by the CLIP text encoder~\cite{clip}, as in the frame embedding extraction. 
Given a text input $T$, all words in the text are tokenized and enclosed with special tokens [\texttt{SOS}] and [\texttt{EOS}], indicating the start and the end of the sentence, respectively. 
This complete sequence is then passed through the CLIP text encoder.
Then, the output corresponding to the [\texttt{EOS}] token is regarded as the text embedding, $\mathbf{t} \in \mathbb{R}^{D}$, capturing the overall semantics of the input text.

\noindent\textbf{Audio Embeddings:} We employ AST~\cite{ast} as our audio encoder following prior work~\cite{ast, tefal}. The audio input $A$ from the video $V$ is first converted into a Mel-spectrogram, capturing key audio features across time and frequency. 
AST processes this spectrogram as a sequence of $N_a$ patches, using transformer layers to generate a detailed set of audio embeddings.
However, since audio signals are much more densely sampled compared to video frames, directly fusing all $N_a$ audio embeddings with frame embeddings would be computationally expensive. 
To address this, we employ an additional query-based transformer~(\ie, audio resampler)~\cite{perceiver, detr, flamingo} that uses a cross-attention mechanism with $M$ learnable query embeddings. 
This design reduces the number of audio embeddings to a fixed length of $M$, $\mathbf{a}=[\mathbf{a}^1, \mathbf{a}^2, \dots, \mathbf{a}^{M}] \in \mathbb{R}^{M \times D}$, maintaining essential audio information while significantly lowering computational load for subsequent fusion with visual modalities. We freeze the pretrained AST parameters during training, avoiding the high computational cost of fine-tuning. Architectural details are provided in the supplementary material.

\subsection{Gated Fusion Transformer}
\label{sec:gate}

To effectively fuse the audio embeddings $\mathbf{a}$ and frame embedding $\mathbf{f}$, the gated fusion transformer is designed to dynamically adjust the contribution of audio based on its relevance to the video. This transformer achieves adaptive fusion by using gating scores that modulate the impact of audio features on the video representation, ensuring that irrelevant audio signals have a reduced influence, thus preserving the integrity of the visual content.

The gated fusion transformer consists of $L$ layers, each containing a gated fusion block that applies gating scores to control the fusion process. Within each block, Multi-Head Attention (MHA) and a Feed-Forward Network (FFN) blend the audio and frame embeddings, with the adaptive gating function providing scores ($g_{\textrm{mha}}$ and $g_{\textrm{ffn}}$) to regulate their influence. These gating scores control the contribution of audio embeddings, enabling the transformer to integrate complementary audio cues when relevant while minimizing the effect of potentially noisy audio. This is followed by Multi-Head Self-Attention (MHSA) and a Feed-Forward Network (FFN) to refine the representation.

When the gating scores are high, audio cues are emphasized, enhancing the representation by capturing complementary audio details. Conversely, low gating scores prioritize the robustness of the visual content against irrelevant audio interference. This selective fusion leads to a more context-sensitive and discriminative video representation, which improves video-text retrieval performance. In the following, we provide architectural details of the two core components of the gated fusion transformer: the gated fusion block and the gating function.

\subsubsection{Gated Fusion Block}
\label{sec:fusion}
The gated fusion block takes the audio embeddings $\mathbf{a}$ and the frame embeddings $\mathbf{f}$ as inputs and outputs video representation through the fusion process followed by the refining process.
Specifically, with a series of $L$ layers, the frame embeddings $\mathbf{f}$ evolve into refined $\mathbf{f}^{(L)}$ by capturing cross-modal interactions with the audio embeddings $\mathbf{a}$, where $\mathbf{f}$ directly from the CLIP image encoder is regarded as initial input $\mathbf{f}^{(0)}$.
The refined $\mathbf{f}^{(L)}$ serves as the final video representation $\mathbf{v}=[\mathbf{v}^1, \mathbf{v}^2, \dots, \mathbf{v}^N]\in\mathbb{R}^{N\times D}$.

In the fusion process, $\mathbf{f}^{(l-1)}$ is fused with $\mathbf{a}$ by MHA with a residual connection.
The output is modulated by a gating score $g_{\textrm{mha}}$ generated by the gating function~(\cref{sec:gating_function}) to determine the contribution of the audio embedding in the fusion.
Then, the output is fed into an FFN with a residual connection, which is also modulated by a gating score $g_{\textrm{ffn}}$, ensuring selective enhancement based on the gating mechanism.
The fusion process is formulated as follows:
\begin{equation}
    \begin{aligned}
    \mathbf{z}^{(l)} &=  g_{\textrm{mha}}^{(l)} \cdot \textrm{MHA}(\textrm{LN}(\mathbf{f}^{(l-1)}), \textrm{LN}(\mathbf{a})) +\mathbf{f}^{(l-1)} , \\
    \bar{\mathbf{z}}^{(l)} &= g_{\textrm{ffn}}^{(l)} \cdot \textrm{FFN}_1(\textrm{LN}(\mathbf{z}^{(l)})) + \mathbf{z}^{(l)},
    \label{eq:fusion_process}
    \end{aligned}
\end{equation}
where LN denotes the layer normalization.
MHA involves $n$-head attention operations; it is formulated by
\begin{equation}
\begin{aligned}
    \textrm{MHA}(\mathbf{X},\mathbf{Y}) &= [\textrm{Attn}^1(\mathbf{X},\mathbf{Y}), ..., \textrm{Attn}^n(\mathbf{X},\mathbf{Y})]\mathbf{W}_o,\\ 
    \textrm{Attn}^i(\mathbf{X},\mathbf{Y}) &= \text{softmax}\left( \frac{(\mathbf{X}\mathbf{W}_{q}^{i}) (\mathbf{Y}\mathbf{W}_k^{i})^\top}{\sqrt{D_h}} \right) (\mathbf{Y}\mathbf{W}_v^{i})
    \label{eq:MHA}
\end{aligned}
\end{equation}
where $[\cdot,\cdot]$ denotes concatenation, and $D_h$ is set to $D/n$. 
$\mathbf{W}_{q}^{i}, \mathbf{W}_{k}^{i}, \mathbf{W}_{v}^{i}\in\mathbb{R}^{D\times D_h}$ and $\mathbf{W}_{o}\in\mathbb{R}^{D\times D}$ are linear projection matrices.

The fused representation $\bar{\mathbf{z}}^{(l)}$ undergoes the refining process to enhance the overall video representation.
Specifically, $\bar{\mathbf{z}}^{(l)}$ is fed into a Multi-Head Self-Attention (MHSA) module with a residual connection.
Then, the second FFN with a residual connection is applied to the output of MHSA to produce the refined frame embeddings $\mathbf{f}^{(l)}$. 
The refining process is formulated as follows:
\begin{equation}
\begin{aligned}
    \Tilde{\mathbf{z}}^{(l)} &= \textrm{MHSA}(\textrm{LN}(\bar{\mathbf{z}}^{(l)})) + \bar{\mathbf{z}}^{(l)} , \\
    \mathbf{f}^{(l)} &= \textrm{FFN}_2(\textrm{LN}(\Tilde{\mathbf{z}}^{(l)})) + \Tilde{\mathbf{z}}^{(l)}.
    \label{eq:refining_process}
\end{aligned}
\end{equation}
Similar to MHA, MHSA involves $n$-head self-attention operations, where $\textrm{MHSA}(\textrm{X})=\textrm{MHSA}(\textrm{X},\textrm{X})$.

\subsubsection{Gating Function}
\label{sec:gating_function}
The gating function, $G(\mathbf{a}, \mathbf{f}^{(l-1)})$ at each layer $l$ produces two gating scores, $g_\text{mha}^{(l)}$ and $g_\text{ffn}^{(l)}$, that modulate the outputs of MHA and FFN$_1$ in the gated fusion block, respectively. 
To this end, the audio embeddings $\mathbf{a}$ and the frame embeddings $\mathbf{f}^{(l-1)}$ are first aggregated using average pooling, resulting in $\bar{\mathbf{a}} \in \mathbb{R}^D$ and $\bar{\mathbf{f}}^{(l-1)} \in \mathbb{R}^D$, respectively. 
These aggregated embeddings are then concatenated to form a joint representation, $\mathbf{u}^{(l)} = [\bar{\mathbf{a}}; \bar{\mathbf{f}}^{(l-1)}] \in \mathbb{R}^{2D}$.
The joint embedding $\mathbf{u}^{(l)}$ is passed through two distinct Multi-Layer Perceptrons (MLPs) followed by a non-linear function to compute the gating scores for the outputs of MHA and FFN$_1$. 
The gating function, $G(\mathbf{a}, \mathbf{f}^{(l-1)})$ is formulated as follows:
\begin{equation}
\begin{aligned}
    [g_{\text{mha}}^{(l)},\ g_{\text{ffn}}^{(l)}] &= G(\mathbf{a}, \mathbf{f}^{(l-1)}) \\
    &=\sigma\Big[\left( \textrm{MLP}_{\textrm{mha}}(\mathbf{u}^{(l-1)})\right), \left( \textrm{MLP}_{\textrm{ffn}}(\mathbf{u}^{(l-1)})\right)\Big],
\label{eq:gates}    
\end{aligned}
\end{equation} 
where $\sigma$ is a non-linear function (\ie, tanh).

\subsection{Adaptive Margin-based Contrastive Learning}
\label{sec:margin_loss}

The output of the gated fusion transformer, video representation $\mathbf{v} \in \mathbb{R}^{N \times D}$, is aligned with text embedding $\mathbf{t} \in \mathbb{R}^{D}$. 
This video-text alignment has been, in general, achieved by the cross-modal contrastive learning~\cite{clip,clip4clip,xpool,xclip}, which maximizes similarity scores (e.g., cosine similarity) of positive pairs and minimizes the those of negative pairs.

To further enhance the discriminative ability of contrastive learning, we introduce an adaptive margin-based contrastive loss that dynamically adjusts margins for each negative pair based on their intra-modal semantic similarities. The key idea of our method is that semantic relationships within each visual and textual modality provide implicit cues for determining appropriate margins. For instance, if two videos are visually similar or if two texts have high textual similarity, their cross-modal counterparts are also likely to have some level of semantic relatedness. 
In contrast to fixed-margin contrastive loss~\cite{kim2020proxy,deng2019arcface, kim2024efficient}, which applies the same margin to all negative pairs, our method accounts for varying degrees of semantic similarity between negative pairs. This adaptability enables the model to learn discriminative features while maintaining strong generalization by considering the inherent relationships between semantically similar pairs.

Following this nature, given a batch with $B$ video-text pairs $\mathcal{B} =\big\{(V_i,T_i)\big\}_{i=1}^{B}$, we define the adaptive margin $m_{ij}$ for each negative pair ($V_i$,$T_j$) according to the average of their intra-modal similarities in both the visual and textual modalities. 
First of all, the frame embedding $\mathbf{f}_i$ is average pooled to obtain holistic embedding of the visual modality denoted by $\bar{\mathbf{f}_i}$.
Let $c_{ij}^{v}$ denotes the cosine similarity between $\bar{\mathbf{f}_i}$ and $\bar{\mathbf{f}_j}$, and $c_{ij}^{t}$ denotes the cosine similarity between $\mathbf{t}_i$ and $\mathbf{t}_j$. The adaptive margin is obtained by:
\begin{equation}
    m_{ij} = \min\left(\lambda \left( 1 - \frac{c_{ij}^{v} + c_{ij}^{t}}{2} \right),\ \delta \right),
    \label{eq:margin}
\end{equation}
where $\lambda$ is a scaling factor, and $\delta$ is the maximum margin to prevent excessively large margins. The operation $\min(\cdot,\cdot)$ ensures that the margin does not exceed $\delta$.
The adaptive margin $m_{ij}$ is then added to the similarity score of the negative pair in the contrastive loss function as follows:
\begin{equation}
\begin{aligned}
    \mathcal{L}(\mathcal{B}) = &-\sum_{i=1}^{B}\Bigg(\log\frac{e^{s_{ii}/\tau }}{e^{s_{ii}/\tau}+\sum_{j=1, j\neq i}^{B}e^{(s_{ij}+m_{ij})/\tau }} \Bigg) \\
    &-\sum_{i=1}^{B}\Bigg(\log\frac{e^{s_{ii}/\tau }}{e^{s_{ii}/\tau}+\sum_{j=1, j\neq i}^{B}e^{(s_{ji}+m_{ji})/\tau }} \Bigg).
    \label{eq:Margin_loss}
\end{aligned}
\end{equation}
By incorporating the adaptive margin into the contrastive loss, our method promotes a more discriminative and generalizable cross-modal embedding space that considers inherent intra-modality relationships, which leads to improved retrieval performance.

\begin{table*}[!t]
\fontsize{8}{9}\selectfont
\centering
\begin{tabularx}{0.98\textwidth}
    {
      p{0.18\textwidth}
      >{\raggedright\arraybackslash}p{0.07\textwidth}
      >{\centering\arraybackslash}X
      >{\centering\arraybackslash}X
      >{\centering\arraybackslash}X
      >{\centering\arraybackslash}X
      >{\centering\arraybackslash}X
      >{\centering\arraybackslash}X 
      >{\centering\arraybackslash}X
      }
     \toprule
    {\multirow{1}{*}[-4.5mm]{\textbf{Methods}}}& \multirow{1}{*}[-4.5mm]{\textbf{Modality}}&
    \multicolumn{3}{c}{\textbf{Text-to-Video Retrieval}} & \multicolumn{3}{c}{\textbf{Video-to-Text Retrieval}} & \multirow{2}{*}{RSum} \\ [-0.3ex]  \cmidrule(lr){3-5}  
    \cmidrule(lr){6-8} & & R@1 & R@5 & R@10 &  R@1 & R@5 & R@10 & \\  [-0.4ex] \midrule
    \multicolumn{9}{l}{\textit{\textbf{CLIP ViT-B/32}}} \\ [-0.3ex]\midrule
    CLIP4Clip$_{meanP}$~\cite{clip4clip} & V+T &43.1 & 70.4 & 80.8 & 43.1 & 70.5 & 81.2 & 389.1 \\
    CLIP4Clip$_{seqTransf}$~\cite{clip4clip}& V+T & 44.5 & 71.4 & 81.6 & 42.7 & 70.9 & 80.6 & 391.7 \\
    ECLIPSE~\cite{eclipse} & A+V+T & 44.9 & 71.3 & 81.6 & 44.7 & 71.9 & 82.8 & 397.2 \\
    X-Pool~\cite{xpool} & V+T & 46.9 & 72.8 & 82.2 & 44.4 & 73.3 & 84.0 & 403.6 \\
    TS2-Net~\cite{ts2} & V+T & 47.0 & 74.5 & {83.8} & 45.3 & 74.1 & 83.7 & 408.4 \\
    UATVR~\cite{uatvr} & V+T & 47.5 & 73.9 & 83.5 & 46.9 & 73.8 & {83.8} & 409.4 \\
    ProST~\cite{prost} & V+T & 48.2 & 74.6 & 83.4 & 46.3 & 74.2 & 83.2 & 409.9 \\
    UCoFiA$^\dagger$~\cite{ucofia}  & V+T &  {49.4} & 72.1 & 83.5 & {47.1} & 74.3 & 83.0 & 409.4 \\
    TEFAL~\cite{tefal} & A+V+T & {49.4} & \textbf{75.9} & \textbf{83.9} & {47.1} & {75.1} & \textbf{84.9} & 416.3 \\
    \ccol  AVIGATE (Ours) & \ccol A+V+T &\ccol  \textbf{50.2} & \ccol {74.3}  &\ccol  83.2 &\ccol \textbf{49.7} &  \ccol  \textbf{75.3}   & \ccol 83.7 & \ccol \textbf{416.4}    \\
    \midrule
    \multicolumn{9}{l}{\textit{\textbf{CLIP ViT-B/16}}} \\ [-0.4ex] \midrule
    X-Pool~\cite{xpool} & V+T & 48.2 & 73.7 & 82.6 & 46.4 & 73.9 & 84.1 & 408.9 \\
    TS2-Net~\cite{ts2} & V+T & 49.4 & 75.6 & 85.3 & 46.6 & 75.9 & 84.9 & 417.7 \\
    ProST~\cite{prost} & V+T & 49.5 & 75.0 & 84.0 & 48.0 & 75.9 & 85.2 & 417.6 \\
    UCoFiA$^\dagger$~\cite{ucofia}  & V+T &  49.8 & 74.6 & 83.5 & {49.1} & {77.0} & 83.8 & 417.8 \\
    TEFAL~\cite{tefal} & A+V+T & 49.9 & 76.2 & 84.4  & - & - & - & - \\
    UATVR~\cite{uatvr} & V+T & {50.8} & {76.3} & \textbf{85.5} & 48.1 & 76.3 & {85.4} & 422.4 \\ 
    \ccol  AVIGATE (Ours) & \ccol A+V+T &\ccol  \textbf{52.1} & \ccol \textbf{76.4} &\ccol  {85.2} &\ccol \textbf{51.2} &  \ccol  \textbf{77.9}   & \ccol \textbf{86.2} & \ccol \textbf{429.0}    \\
\bottomrule 
\end{tabularx}
\vspace{-1.5mm}
\caption{Text-to-video and video-to-text retrieval results on the MSR-VTT 9k split. 
$^\dagger$ denotes the use of post-processing techniques.
}
\label{tab:main_results}
\vspace{-4.5mm}
\end{table*}

To compute a similarity score, rather than simply leveraging cosine similarity, we employ an enhanced method that captures both global and local relationships between video and text representations. This combination captures complementary information, enhancing the retrieval performance as observed in the previous arts~\cite{xclip,ucofia,uatvr,prost}. 
First, for global alignment, we derive a comprehensive video representation by applying average pooling to the final video embedding, $\mathbf{v}=[\mathbf{v}^1, \mathbf{v}^2, \dots, \mathbf{v}^N]\in\mathbb{R}^{N\times D}$, where $N$ represents the number of frame segments. This pooled representation,  $\mathbf{v}_g = \sum_{i=1}^{N} \mathbf{v}^i / N \in \mathbb{R}^{D}$, captures the overall content of the video. The video-text similarity score $s_g$ between the global representation of the video $\mathbf{v}_g$ and the text embedding $\mathbf{t} \in \mathbb{R}^{D}$ is then computed as:
\begin{equation}
    s_g = \frac{\mathbf{v}_g^\top \mathbf{t}}{|| \mathbf{v}_g ||_2 || \mathbf{t} ||_2}.
\end{equation}
Next, for local alignment, we match each frame segment embedding $\mathbf{v}^i \in \mathbb{R}^{D}$ with the text embedding $\mathbf{t}$, computing frame-text similarities $s^i = \frac{\mathbf{v}^{i\top} \mathbf{t}}{|| \mathbf{v}^i ||_2 || \mathbf{t} ||_2}$. These frame-text similarities are aggregated using the log-sum-exp (LSE) function:
\begin{equation}
    s_l = \underset{{i\in[1,N]}}{\text{LSE}}(s^i) = \log\left( \sum_{i=1}^{N} e^{\alpha s^i} \right),
    \label{eq:local_alignment}
\end{equation}
where $\alpha$ is a scaling parameter that controls the smoothness of the maximum operation.
The final similarity score $s$, which combines both global and local perspectives, is calculated by averaging $s_g$ and $s_l$:
\begin{equation}
    s = \frac{s_g + s_l}{2}. 
    \label{eq:final_score}
\end{equation}
The final similarity score $s$ is utilized for the adaptive margin-based contrastive loss in~\cref{eq:Margin_loss}.
This multi-grained alignment scheme enables the model to capture both holistic and fine-grained details, thereby enhancing video-text retrieval accuracy.


\section{Experiments}
\label{sec:experiments}
In this section, we present a comprehensive evaluation of the proposed framework, \textit{AVIGATE}. 
We first describe the experimental setup, including the datasets, evaluation metric, and implementation details in~\cref{sec:setup}. 
We then provide quantitative results to show the effectiveness of AVIGATE compared to state-of-the-art methods in~\cref{sec:quan}. 
Next, we include the qualitative results of AVIGATE in~\cref{sec:qual}. 
Moreover, we conduct ablation studies to analyze the impact of components of AVIGATE in~\cref{sec:ablation}.
Lastly, we demonstrate the efficiency of AVIGATE compared to previous arts by analyzing the computational costs in~\cref{sec:complexity}.
\subsection{Experimental Setup}
\label{sec:setup}
\noindent\textbf{Datasets:}
On \bs{three} public benchmarks, MSR-VTT~\cite{msrvtt}, VATEX~\cite{vatex} \bs{and Charades~\cite{charades}}, we evaluate and compare the performance of our method against previous methods.
MSR-VTT, the most common dataset for video-text retrieval, contains 10,000 videos collected from the web, including audio signals. Each video is 10 to 32 seconds long and has 20 corresponding text descriptions. Following the data split from~\cite{mmt,clip4clip}, we train AVIGATE on 9,000 videos (180,000 video-text pairs) and evaluate it on 1,000 selected video-text pairs. Out of the 10,000 videos, 9,582 have audio, which we utilize in our method.
VATEX contains 34,991 videos with multiple text descriptions for each video.  We follow the split protocol of~\cite{hgr} with 25,991 videos for training, 1,500 videos for validation, and 1,500 for testing, respectively.
\bs{Charades consists of 9,848 videos, each paired with a single textual description. 
We follow the split protocol of~\cite{tefal, eclipse}.}

\noindent \textbf{Evaluation Metric:} 
We employ the standard metric of recall at $K$ (R@$K$, with $K$ = 1, 5, 10) to evaluate retrieval performance for both text-to-video and video-to-text retrieval tasks. In both cases, samples are ranked based on their similarities to the query (in~\cref{eq:final_score}). 
We also report the RSUM, which is the sum of all R@$K$ metrics.
Retrieval is considered successful if at least one relevant item appears within the Top-K positions.

\begin{figure*}[t!]
    \vspace{-1mm}
    \centering
    \includegraphics[width=0.9\linewidth]{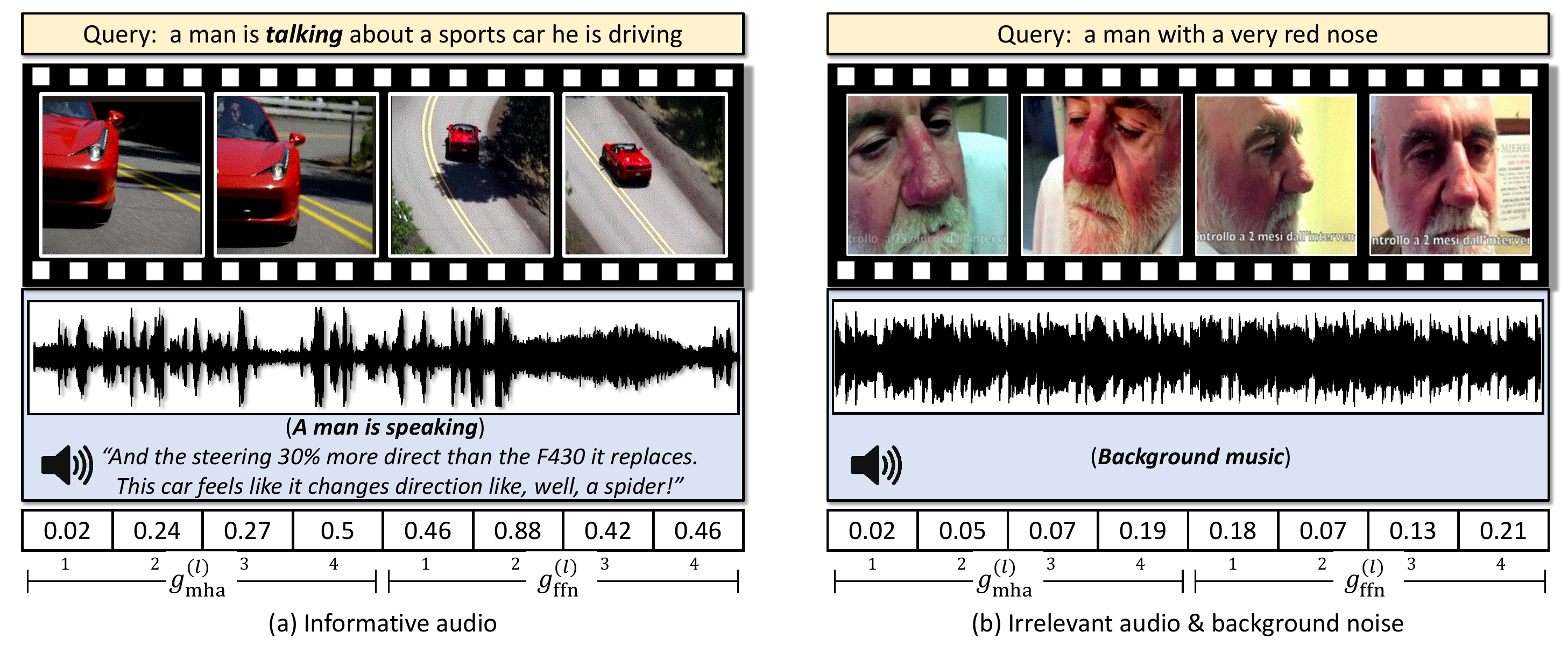}
    \vspace{-3.5mm}
    \caption{ Top-1 text-to-video retrieval results of our method on MSR-VTT, where they are true matches. 
    $g_{mha}^{(l)}$ and $g_{ffn}^{(l)}$ denote the gating scores for $l$-th layers of the gated fusion transformer. 
    The audio provides informative cues for accurate retrieval, where ``a man is talking'' in the query text is not visible (a). The irrelevant audio is filtered by the gated fusion transformer, leading to an accurate retrieval result (b).   
    }
    \label{fig: qual}
    \vspace{-4.5mm}
\end{figure*}

\noindent\textbf{Implementation Details:}
We adopt the pre-trained CLIP models from OpenAI~\cite{clip} with two different sizes of image encoder, \ie, ViT-B/32 and ViT-B/16. Moreover, we adopt AST pre-trained on ImageNet~\cite{imagenet} and AudioSet~\cite{audioset} with DeiT~\cite{DeiT} backbone.
Note that we set the audio input as the zero vector for the videos that do not include audio signals. 
\bs{More details on implementation are presented in the supplementary material. (See Sec. B)}

\subsection{Quantitative Results}
\label{sec:quan}
\bs{We compare AVIGATE with previous video-text retrieval methods on the MSR-VTT 9k split~\cite{msrvtt}, VATEX~\cite{vatex} and Charades~\cite{charades} with results presented in Table~\ref{tab:main_results}, Table~\ref{tab:vatex} and Table~\ref{tab:chardes}, respectively.}
Each table shows the modalities used by the methods for video-text retrieval, indicated as visual (V), textual (T), and audio (A), with results reported separately for each CLIP ViT backbone size.

\begin{table}[!t]
\vspace{2mm}
\centering
\fontsize{8}{8.7}\selectfont
\scalebox{0.95}{
\begin{tabularx}{0.499\textwidth}
    {
      p{0.13\textwidth}
      >{\raggedright\arraybackslash}p{0.07\textwidth}
      >{\centering\arraybackslash}X
      >{\centering\arraybackslash}X
      >{\centering\arraybackslash}X
      }
     \toprule
    {\multirow{1}{*}[-4.5mm]{\textbf{Methods}}}& \multirow{1}{*}[-4.5mm]{\textbf{Modality}}&
    \multicolumn{3}{c}{\textbf{Text-to-Video Retrieval}} \\ [-0.3ex]  \cmidrule(lr){3-5}  
     & & R@1 & R@5 & R@10   \\  [-0.4ex] \midrule
    \multicolumn{5}{l}{\textit{\textbf{CLIP ViT-B/32}}} \\ [-0.3ex]\midrule
    ECLIPSE~\cite{eclipse} & A+V+T & 57.8 & 88.4 & 94.3 \\
    X-Pool~\cite{xpool} & V+T & 60.0 & 90.0 & 95.0 \\
    ProST~\cite{prost} & V+T & 60.6 & 90.5 & 95.4 \\ 
    TEFAL~\cite{tefal} & A+V+T & 61.0 & 90.0 & 95.0 \\
    UCoFiA$^\dagger$~\cite{ucofia} & V+T & 61.1 & 90.5 & - \\
    UATVR~\cite{uatvr} & V+T & {61.3} & \textbf{91.0} & \textbf{95.6} \\
    \ccol  AVIGATE (Ours) & \ccol A+V+T &\ccol  \textbf{63.1} & \ccol {90.7 } &\ccol  {95.5} \\
    \midrule
    \multicolumn{5}{l}{\textit{\textbf{CLIP ViT-B/16}}} \\ [-0.4ex] \midrule
    X-Pool~\cite{xpool} & V+T & 62.6 & 91.7 & 96.0 \\
    ProST~\cite{prost} & V+T & 64.0 & 92.2 & 96.3 \\
    UATVR~\cite{uatvr} & V+T & {64.5} & {92.6} & \textbf{96.8} \\
    \ccol  AVIGATE (Ours) & \ccol A+V+T &\ccol  \textbf{67.5} & \ccol \textbf{93.2 } &\ccol  {96.7} \\
     
\bottomrule 
\end{tabularx}
}
\vspace{-1.5mm}
\caption{Text-to-video retrieval results on VATEX. 
}
\label{tab:vatex}
\vspace{-4.5mm}
\end{table}

These tables demonstrate that AVIGATE outperforms all previous methods in R@1 across all datasets and CLIP ViT backbone sizes.
Specifically, on MSR-VTT, AVIGATE achieves improvements of 0.8\%p and 2.6\%p over TEFAL~\cite{tefal} (w/ audio SOTA) in text-to-video retrieval and video-to-text retrieval, respectively.
Ours surpasses UATVR~\cite{uatvr} (w/o audio, but SOTA) by a large margin, 1.8\%p in R@1 on VATEX.
\bs{On Charades, AVIGATE also outperforms TEFAL by 0.8\%p in R@1.}
The improvements of AVIGATE become more distinctive when a larger backbone is used.
Specifically, on MSR-VTT, AVIGATE achieves a 1.3\%p improvement in text-to-video retrieval and a 3.1\%p in video-to-text retrieval while achieving a 3.0\%p improvement in text-to-video retrieval on VATEX.
Furthermore, our method demonstrates significant improvements in RSum, with 6.6\%p improvements on MSR-VTT.
The superior performance of AVIGATE is attributed to its ability to selectively integrate informative audio cues while dynamically adjusting the semantic alignment between video and text, leading to a more discriminative and generalizable cross-modal representation.
\begin{table}[!t]
\vspace{2mm}
\centering
\fontsize{8}{8.7}\selectfont
\scalebox{0.95}{
\begin{tabularx}{0.499\textwidth}
    {
      p{0.13\textwidth}
      >{\raggedright\arraybackslash}p{0.07\textwidth}
      >{\centering\arraybackslash}X
      >{\centering\arraybackslash}X
      >{\centering\arraybackslash}X
      }
     \toprule
    {\multirow{1}{*}[-4.5mm]{\textbf{Methods}}}& \multirow{1}{*}[-4.5mm]{\textbf{Modality}}&
    \multicolumn{3}{c}{\textbf{Text-to-Video Retrieval}} \\ [-0.3ex]  \cmidrule(lr){3-5}  
     & & R@1 & R@5 & R@10   \\  [-0.4ex] \midrule
    \multicolumn{5}{l}{\textit{\textbf{CLIP ViT-B/32}}} \\ [-0.3ex]\midrule
    X-Pool~\cite{xpool} & V+T & 16.1 & 35.2 & 44.9 \\
    TEFAL~\cite{tefal} & A+V+T & 18.2 & 37.3 & 48.6 \\
    \ccol  AVIGATE (Ours) & \ccol A+V+T &\ccol  \textbf{18.8} & \ccol \textbf{40.0} &\ccol  \textbf{51.8} \\
    \midrule
    \multicolumn{5}{l}{\textit{\textbf{CLIP ViT-B/16}}} \\ [-0.4ex] \midrule
    \ccol  AVIGATE (Ours) & \ccol A+V+T &\ccol  \textbf{24.1} & \ccol \textbf{48.5} &\ccol  \textbf{61.3} \\
     
\bottomrule 
\end{tabularx}
}
\vspace{-1.5mm}
\caption{\bs{Text-to-video retrieval results on Charades.} 
}
\label{tab:chardes}
\vspace{-4.5mm}
\end{table}

\subsection{Qualitative Results}
\label{sec:qual}
Figure~\ref{fig: qual} illustrates the Top-1 retrieved video results from our method for the given text query, including the paired audio signal and layer-wise gating scores.
For the query in Figure~\ref{fig: qual}(a), our model retrieves the correct video, supported by the informative audio cues that match the content of the query. 
In this example, the audio provides valuable information, reflected by relatively high gating scores ($g_{\text{mha}}^{(l)}$ and $g_{\text{ffn}}^{(l)}$) except $g_{\text{mha}}^{(1)}$ across layers, emphasizing the contribution of the audio in the fusion.
Conversely, for the query in Figure~\ref{fig: qual}(b), the successfully retrieved video includes background music that is irrelevant content.
The gating mechanism responds accordingly, assigning low gating scores to suppress the influence of irrelevant audio signals. 
This behavior shows that the gated fusion transformer successfully filters out irrelevant audio while using the multi-modal nature of videos only when the audio contributes positively. 

\subsection{Ablation Studies}
\label{sec:ablation}
We evaluate the effectiveness of the proposed components in AVIGATE through comprehensive experiments.
For the ablation study, we report text-to-video retrieval results on the MSR-VTT dataset~\cite{msrvtt} with CLIP ViT-B/32~\cite{clip}. 

\noindent \textbf{Components of AVIGATE:}
In Table~\ref{tab:ablation_components}, we present the results of the ablation study on the main components of our model.
We first build a baseline that includes only the CLIP encoders for video and text embeddings, which is learned by a conventional cross-modal contrastive loss.
Then, the baseline is extended by introducing a simplified fusion transformer without the gating mechanism from our gated fusion transformer to exploit audio modality (Baseline + Audio).
The extended baseline improves overall performance, particularly with a notable 1.7\%p improvement in R@1, demonstrating that audio contributes positively to text-to-video retrieval.
Subsequently, introducing our main components, including the gated fusion transformer and adaptive margin, consistently improves the extended baseline in all metrics when each component is applied individually.
This observation highlights the effectiveness of the gated fusion transformer, which dynamically adjusts the contribution of the audio, and the benefit of contrastive learning with adaptive margin.
The full configuration of our method achieves the best performance in R@1 at 50.2\% while demonstrating consistent improvements and highlighting the complementary synergy between the gated fusion transformer and the adaptive margin-based contrastive loss.
It is worth noting that employing a fixed margin ($m_{ij}$ in Eq.~\cref{eq:margin} is fixed as 0.1) with the gated fusion transformer achieves a moderate performance gain in R@1, but the most significant improvement is presented when the margin is adaptively employed.

\begin{table}[!t]
\setlength{\tabcolsep}{4pt}
\fontsize{8}{8.7}\selectfont
\centering
\scalebox{0.95}{
\begin{tabularx}{0.5\textwidth}
    {
      p{0.12\textwidth}
      >{\centering\arraybackslash}p{0.04\textwidth}
      >{\centering\arraybackslash}p{0.04\textwidth}
      >{\centering\arraybackslash}p{0.063\textwidth}
      >{\centering\arraybackslash}X
      >{\centering\arraybackslash}X
      >{\centering\arraybackslash}X
      }
     \toprule
    {\multirow{1}{*}[-4.7mm]{\textbf{Methods}}}& \multirow{1}{*}[-4.7mm]{\textbf{Gate}}&
    \multicolumn{2}{c}{\textbf{Margin}} &
    \multicolumn{3}{c}{\textbf{Text-to-Video Retrieval}} \\ [-0.3ex] \cmidrule(lr){3-4} \cmidrule(lr){5-7}  
     & & \textbf{Fixed} & \textbf{Adaptive} & R@1 & R@5 & R@10   \\  [-0.4ex] \midrule
    Baseline & \xmark & \xmark & \xmark & 45.4 & 72.2 & 81.6 \\
    ~~+ Audio & \xmark & \xmark & \xmark & 47.1& 73.4 & 81.9 \\ \midrule
    Adaptive Margin & \xmark & \xmark & \cmark & 48.9 & 74.8& 83.7\\
    Gated Fusion & \cmark & \xmark & \xmark & 48.0& 75.1 & 83.4\\
    ~~+ Fixed Margin & \cmark & \cmark & \xmark& 49.0 & 74.1 & 83.5\\ 
    \ccol  AVIGATE (Ours) & \ccol \cmark &\ccol \xmark&\ccol  \cmark & \ccol 50.2 &\ccol  74.3 & \ccol 83.2  \\

\bottomrule 
\end{tabularx}
}
\vspace{-1.5mm}
\caption{Ablation studies on key components of our method. 
}
\label{tab:ablation_components}
\vspace{-2.0mm}
\end{table}

\begin{table}[!t]
\setlength{\tabcolsep}{2pt}
\fontsize{8}{8.8}\selectfont
\centering
\scalebox{0.95}{
\begin{tabularx}{0.499\textwidth}
    {
      >{\centering\arraybackslash}p{0.13\textwidth}
      >{\centering\arraybackslash}X
      >{\centering\arraybackslash}X
      >{\centering\arraybackslash}X
      }
     \toprule
    \multirow{1}{*}[-4.5mm]{\textbf{Ablated Setting}} & \multicolumn{3}{c}{\textbf{Text-to-Video Retrieval}} \\ [-0.3ex] \cmidrule(lr){2-4}  
    &  R@1 & R@5 & R@10   \\  [-0.4ex] \midrule
    \multicolumn{4}{l}{\textit{\textbf{Granularity of alignment scheme}}} \\ [-0.3ex]\midrule
    Global only & 46.1 & 73.4 & 82.4 \\
    Local only & 48.5 & 73.7 & 82.8 \\
    \ccol Global-Local   & \ccol 50.2 &\ccol  74.3 & \ccol 83.2  \\

\bottomrule 
\end{tabularx}
}
\vspace{-1.5mm}
\caption{Ablation studies on alignment scheme. 
}
\label{tab:ablation_hyp}
\vspace{-4.5mm}
\end{table}
\noindent \textbf{Granularity of Alignment Scheme:}
We analyze alignment schemes in terms of granularity in Table~\ref{tab:ablation_hyp}, demonstrating that the global-local multi-grained alignment scheme improves performance across all metrics due to its ability to capture the overall semantic context and fine details.


\subsection{Computational Cost Analysis}
\label{sec:complexity}

\begin{table}[!t]
\setlength{\tabcolsep}{2pt}
\fontsize{8}{9.2}\selectfont
\centering
\scalebox{0.95}{
\begin{tabularx}{0.499\textwidth}
    {
      p{0.09\textwidth}
      >{\centering\arraybackslash}p{0.12\textwidth}
      >{\centering\arraybackslash}X
      >{\centering\arraybackslash}X
      >{\centering\arraybackslash}p{0.05\textwidth}
      >{\centering\arraybackslash}p{0.05\textwidth}
      >{\centering\arraybackslash}p{0.05\textwidth}
      }
     \toprule
    {\multirow{1}{*}[-4.5mm]{\textbf{Methods}}}& \multirow{1}{*}[-0.5mm]{\textbf{Time}}&
    \multicolumn{2}{c}{\textbf{Performance}} & \multicolumn{3}{c}{\textbf{Latency (ms)}$\downarrow$}  \\ [-0.3ex] \cmidrule(lr){3-4} \cmidrule(lr){5-7}   
     & \multirow{1}{*}[0.5mm]{\textbf{Complexity}}& R@1 & RSum & $t_\text{sim}$ & $t_\text{ex}$ & Total  \\  [-0.4ex] \midrule
    \multicolumn{7}{l}{\textit{\textbf{Modality: Video and Text}}} \\ [-0.3ex]\midrule
    CLIP4Clip~\cite{clip4clip} & $\mathcal{O}$($\mathcal{V}+\mathcal{T}$) & 44.5 & 391.7 & 0.02 & 9.74 & 9.76 \\
    X-Pool~\cite{xpool} & $\mathcal{O}$($\mathcal{V}\mathcal{T}$)& 46.9 & 403.6  & 56.57 & 9.74 & 66.31 \\ [-0.3ex]\midrule
    \multicolumn{7}{l}{\textit{\textbf{Modality: Audio, Video and Text}}} \\ [-0.3ex]\midrule
    TEFAL~\cite{tefal} &$\mathcal{O}$($\mathcal{A}\mathcal{T} +\mathcal{V}\mathcal{T}$) & 49.4 & 416.3  &130.83 & 9.74 &140.57 \\
    AVIGATE & $\mathcal{O}$($\mathcal{A}+\mathcal{V}+\mathcal{T}$) &\textbf{50.2} & \textbf{416.4}& 0.16 &9.74 &9.90 \\
    \bottomrule 
\end{tabularx}
}
\vspace{-1.5mm}
\caption{Analysis of the efficiency of our method compared to the previous arts. 
$t_\textrm{sim}$ and $t_\textrm{ex}$ denote the latency of the similarity calculation and the query embedding extraction, respectively.
Note that $t_\text{ex}$ for all methods are the same since they use the same text encoder. 
The latency is calculated by a single RTX3090 card.}
\label{tab:complexity}
\vspace{-4.5mm}
\end{table}
Let $\mathcal{T}$, $\mathcal{V}$, and $\mathcal{A}$ denote the number of text queries, videos, and audios, respectively, where $\mathcal{A} \leq \mathcal{V}$, as each audio is associated with a corresponding video but not all videos include an audio.
Table~\ref{tab:complexity} compares the computational complexity and latency of AVIGATE with previous methods~\cite{clip4clip,xpool,tefal} to demonstrate the efficiency of our model during testing.
AVIGATE achieves an efficient retrieval process with a time complexity of $\mathcal{O}(\mathcal{A}+\mathcal{V} + \mathcal{T})$, while TEFAL introduces additional computational burdens with complexity of $\mathcal{O}(\mathcal{AT} + \mathcal{VT})$.
We further evaluate the average latency involved in processing a single text query against a large set of pre-extracted video representations for text-to-video retrieval.
Specifically, we utilize the MSR-VTT-9k test split~\cite{msrvtt}, which comprises 1,000 video-text pairs, where the video representations are pre-extracted and stored, eliminating the need for on-the-fly computation during the retrieval process.
When a text query arrives, the system computes similarity scores between the text embedding and the entire set of video representations. 
The total latency encompasses the extraction of the text embedding and calculating similarity scores between the text embedding and all video representations.
AVIGATE is over 14 times faster than TEFAL and over 6 times faster than X-Pool, as they require cross-modal interaction processes repeatedly for all videos with each text query to generate their final representations, resulting in substantial latency overhead.
Meanwhile, ours incurs only a slight additional cost due to our multi-grained alignment compared to CLIP4Clip.

\section{Conclusion}
\label{sec:conclusion}
We have introduced a novel video-text retrieval framework, AVIGATE, that effectively leverages audio cues through a gated attention mechanism. By selectively filtering out uninformative audio signals, AVIGATE enhances the video representation, leading to a more comprehensive understanding of video content. 
Additionally, we proposed an adaptive margin-based contrastive loss to address the inherent positive-negative relationships between visual and textual modalities. 
This loss function facilitates a more discriminative and generalized cross-modal embedding space by dynamically adjusting margins according to intra-modal semantic dissimilarities. 
Extensive experiments on public benchmarks demonstrate that AVIGATE achieves state-of-the-art performance and ensures efficient retrieval.

\bs{\noindent\textbf{Acknowledgement.} This work was supported by the NRF grant (RS-2021-NR059830--30\%) and the IITP grants (RS-2022-II220290--30\%, RS-2024-00509258--30\%, RS-2019-II191906--10\% 
) funded by Ministry of Science and ICT, Korea.}


\clearpage
{\small
\bibliographystyle{ieeenat_fullname}
\bibliography{cvlab_kwak}
}

\clearpage
\renewcommand\thesection{\Alph{section}}
\setcounter{section}{0}

\section{Appendix}
This supplementary material provides additional details of the audio resampler and experimental results, which we could not include in the main paper. 
We first describe the details of the audio resampler that reduces the number of audio embeddings to a fixed length by employing a query-based transformer~\cite{flamingo,perceiver,detr} \bs{and the details of MLP in Eq.~(4) of the main paper} in~\cref{supp_sec:archi}.
\bs{We also provide implementation details of our method in~\cref{supp_sec:detail}.}
We then present further quantitative results, including the effect of post-processing and entire video-text retrieval results on VATEX~\cite{vatex} \bs{and Charades~\cite{charades}} in~\cref{supp_sec:quan}.
Moreover, we conduct additional ablation studies on hyperparameters, such as the layer depth of the gated fusion transformer, the scaling factor, the maximum margin in Eq.~(5) of the main paper and the type of the gate mechanism, and the effect of freezing modality encoders in~\cref{supp_sec:ablation}.
Lastly, we provide more qualitative results, further illustrating the effectiveness of AVIGATE in~\cref{supp_sec:qual}.

\subsection{More Architectural Details}
\label{supp_sec:archi}
\begin{figure}[t!]
    \centering
    \includegraphics[width=0.99\linewidth]{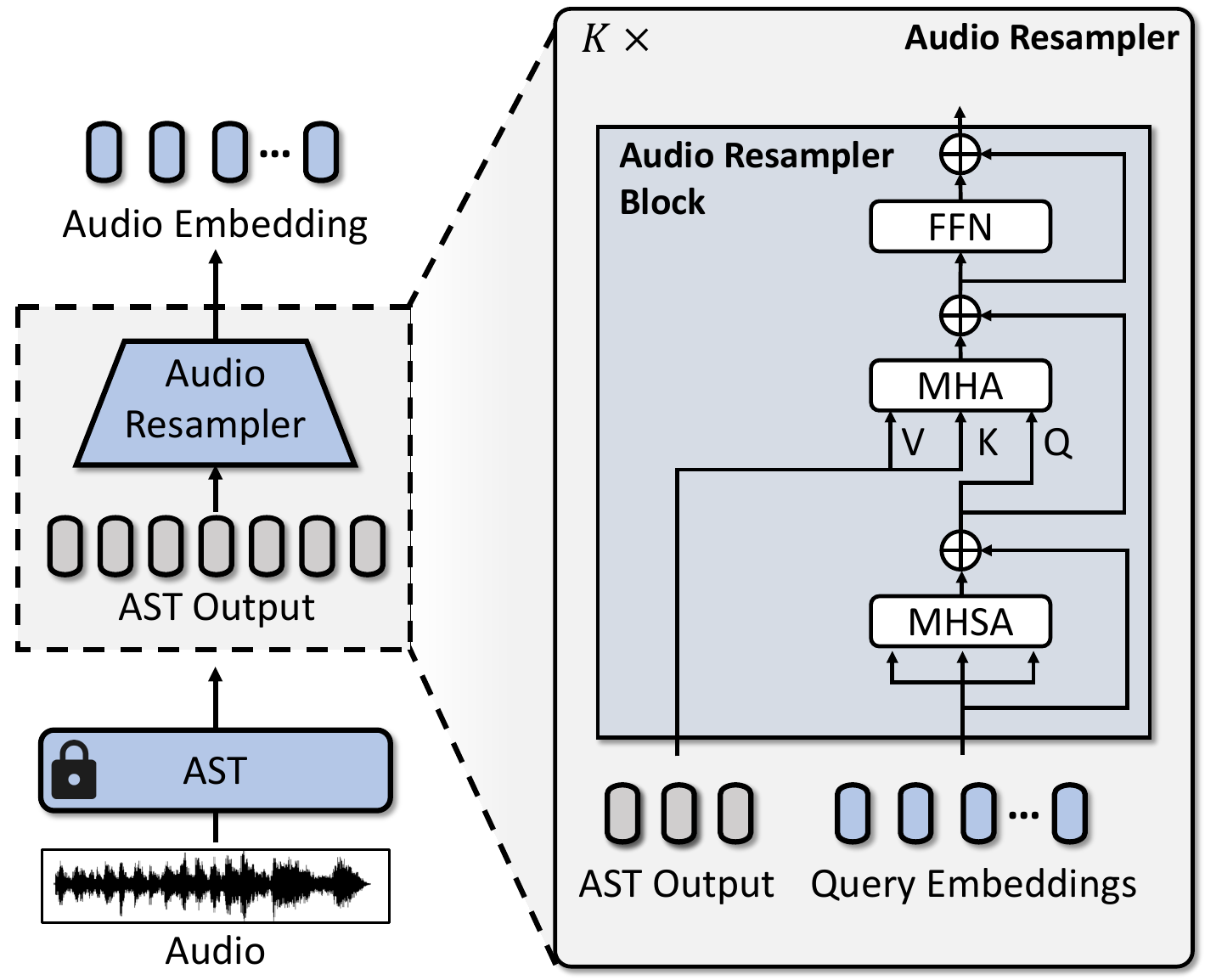}
    \caption{The overall architecture of audio resampler.}
    \label{fig:resampler}
    \vspace{-3mm}
\end{figure}

To efficiently fuse audio embeddings with frame embeddings while reducing computational overhead, we introduce an audio resampler using a query-based transformer framework~\cite{perceiver,detr,flamingo} that utilizes a cross-attention mechanism with $M$ learnable query embeddings. 
Specifically, the audio input is fed into Audio Spectrogram Transformer (AST)~\cite{ast}, and the output is then passed to the audio resampler to reduce the number of audio embeddings to a fixed length of $M$ while preserving essential information.
As shown in Figure~\ref{fig:resampler}, the audio resampler comprises $K$ audio resampler blocks, each with multi-head self-attention (MHSA), multi-head cross-attention (MHA), and a feed-forward network (FFN).
We set $K$ as 4 for default. 

MHSA first allows the learnable query embeddings to interact and capture contextual relationships among themselves, refining their initial representations. This is followed by MHA, where the query embeddings attend to the output of AST, extracting audio embeddings with a fixed length of $M$.
The FFN then processes the audio embeddings to refine them.
This sequence of operations enables the audio resampler to reduce the number of audio embeddings efficiently while preserving critical information, facilitating seamless fusion with the frame embeddings in subsequent stages.

\bs{The MLP in Eq. (4) consists of two layers with dimensions $\mathbb{R}^{2D\times D/2}$ and $\mathbb{R}^{D/2 \times 1}$, using a QuickGELU as the non-linearity between them.
}

\begin{table*}[!t]
    \fontsize{9.5}{10.5}\selectfont
    \setlength{\tabcolsep}{4pt}
    \centering
    \scalebox{0.95}{
    \begin{tabular}{l|ccc}
    \toprule
    Source dataset & MSR-VTT~\cite{msrvtt} & VATEX~\cite{vatex}  & Charades~\cite{charades} \\\midrule
    Image encoder & \multicolumn{3}{c}{2 CLIP-ViTs (B/32 and B/16)} \\ \midrule
    Total epochs & \multicolumn{3}{c}{5}  \\
    Optimizer & \multicolumn{3}{c}{Adam~\cite{Adamsolver}} \\
    Embedding dimension $D$ & \multicolumn{3}{c}{512} \\
    Batch size & 128 & 128 & 64  \\ 
    Max frames & 12 & 12 & 32 \\
    Max words & 32 & 32 & 64 \\
    Resampled audio length & \multicolumn{3}{c}{12} \\
    Depth of Gated Fusion Transformer $L$& \multicolumn{3}{c}{4} \\
    Learning rate for Non-CLIP parameters & $\expnum{1}{4}$ &$\expnum{1}{4}$ & $\expnum{5}{4}$ \\
    Learning rate for CLIP encoders & \multicolumn{3}{c}{$\expnum{1}{7}$} \\
    Temperature $\tau$ in Eq.(6)& \multicolumn{3}{c}{Learnable (After training: 0.01)}  \\
    Maximum margin $\delta$ in Eq.(5) & 0.1 & 0.05 & 0.1 \\
    Scaling factor $\lambda$ in Eq.(5) & \multicolumn{3}{c}{0.2} \\
    Scaling factor $\alpha$ in Eq.(8) & \multicolumn{3}{c}{50} \\
    
    \bottomrule
    \end{tabular}
    }
    \vspace{-2mm}
    \caption{Training configurations of various datasets.}
    \label{tab:config}
    \vspace{-2mm}
\end{table*}

\begin{table*}[!t]
\fontsize{8}{9.2}\selectfont
\centering
\begin{tabularx}{0.98\textwidth}
    {
      p{0.153\textwidth}
      >{\raggedright\arraybackslash}p{0.07\textwidth}
      >{\raggedright\arraybackslash}p{0.075\textwidth}
      >{\raggedright\arraybackslash}p{0.075\textwidth}
      >{\raggedright\arraybackslash}p{0.075\textwidth}
      >{\raggedright\arraybackslash}p{0.075\textwidth}
      >{\raggedright\arraybackslash}p{0.075\textwidth}
      >{\raggedright\arraybackslash}p{0.075\textwidth}
      >{\raggedright\arraybackslash}X
      }
     \toprule
    {\multirow{1}{*}[-4.5mm]{\textbf{Methods}}}& \multirow{1}{*}[-4.5mm]{\textbf{Modality}}&
    \multicolumn{3}{c}{\textbf{Text-to-Video Retrieval}} & \multicolumn{3}{c}{\textbf{Video-to-Text Retrieval}} & \multicolumn{1}{c}{\multirow{2}{*}{RSum}} \\ [-0.3ex]  \cmidrule(lr){3-5}  
    \cmidrule(lr){6-8} & & R@1 & R@5 & R@10 &  R@1 & R@5 & R@10 & \\  [-0.4ex] \midrule
    \multicolumn{9}{l}{\textit{\textbf{CLIP ViT-B/32}}} \\ [-0.3ex]\midrule
    CAMoE~\cite{camoe} & V+T& 44.6 & 72.6 & 81.8 & 45.1 & 72.4 & 83.1 & 399.6 \\
    ~~+DSL& V+T & 47.3 \textcolor{grn}{(+2.7)} & 74.2 \textcolor{grn}{(+1.6)} & 84.5 \textcolor{grn}{(+2.7)} & 49.1 \textcolor{grn}{(+4.0)} & 74.3 \textcolor{grn}{(+1.9)} & 84.3 \textcolor{grn}{(+1.2)} & 413.7 \textcolor{grn}{(+14.1)} \\
    TS2-Net~\cite{ts2} & V+T & 47.0 & 74.5 & {83.8} & 45.3 & 74.1 & 83.7 & 408.4 \\
    ~~+DSL & V+T & 51.1 \textcolor{grn}{(+4.1)} & 76.9 \textcolor{grn}{(+2.4)} & 85.6 \textcolor{grn}{(+1.8)} & - & - & - & - \\
    UATVR~\cite{uatvr} & V+T & 47.5 & 73.9 & 83.5 & 46.9 & 73.8 & {83.8} & 409.4 \\
    ~~+DSL & V+T & 49.8 \textcolor{grn}{(+2.3)} & 76.1 \textcolor{grn}{(+2.2)} & 85.5 \textcolor{grn}{(+2.0)} & 51.1 \textcolor{grn}{(+4.2)} & 74.8 \textcolor{grn}{(+1.0)} & 85.1 \textcolor{grn}{(+1.3)} & 422.4 \textcolor{grn}{(+13.0)} \\
    UCoFiA~\cite{ucofia}  & V+T &  48.2 & 73.3 & 82.3 & - & - & - & - \\
    ~~+SK norm  & V+T &  49.4 \textcolor{grn}{(+1.2)} & 72.1 \textcolor{red}{(-0.9)} & 83.5 \textcolor{grn}{(+1.2)} & {47.1} & 74.3 & 83.0 & 409.4 \\

    \ccol  AVIGATE (Ours) & \ccol A+V+T &\ccol  {50.2} & \ccol {74.3}  &\ccol  83.2 &\ccol {49.7} &  \ccol  {75.3}   & \ccol 83.7 & \ccol {416.4}    \\
    \ccol  ~~+DSL & \ccol A+V+T &\ccol  \textbf{53.9} \textcolor{grn}{(+3.7)} & \ccol \textbf{77.0} \textcolor{grn}{(+2.7)}  &\ccol  \textbf{86.0} \textcolor{grn}{(+2.8)} &\ccol \textbf{53.0} \textcolor{grn}{(+3.3)} &  \ccol  \textbf{78.2} \textcolor{grn}{(+2.9)}   & \ccol \textbf{85.4} \textcolor{grn}{(+1.7)} & \ccol \textbf{433.5} \textcolor{grn}{(+16.9)}     \\    
    \midrule
    \multicolumn{9}{l}{\textit{\textbf{CLIP ViT-B/16}}} \\ [-0.4ex] \midrule
    TS2-Net~\cite{ts2} & V+T & 49.4 & 75.6 & 85.3 & 46.6 & 75.9 & 84.9 & 417.7 \\
    ~~+DSL & V+T & 54.0 \textcolor{grn}{(+4.6)} & 79.3 \textcolor{grn}{(+3.7)} & 87.4 \textcolor{grn}{(+2.1)} & - & - & - & - \\
    TEFAL~\cite{tefal} & A+V+T & 49.9 & 76.2 & 84.4  & - & - & - & - \\
    ~~+DSL+QB-Norm & A+V+T & 52.0 \textcolor{grn}{(+2.1)} & 76.6 \textcolor{grn}{(+0.4)} & 86.1 \textcolor{grn}{(+1.7)}  & - & - & - & - \\
    UATVR~\cite{uatvr} & V+T & {50.8} & {76.3} & 85.5 & 48.1 & 76.3 & {85.4} & 422.4 \\ 
    ~~+DSL & V+T & {53.5} \textcolor{grn}{(+2.7)} & {79.5} \textcolor{grn}{(+3.2)} & 88.1 \textcolor{grn}{(+2.7)} & 54.5 \textcolor{grn}{(+6.4)} & 79.1 \textcolor{grn}{(+2.8)} & \textbf{87.9} \textcolor{grn}{(+2.5)} & 442.6 \textcolor{grn}{(+20.2)} \\ 
    \ccol  AVIGATE (Ours) & \ccol A+V+T &\ccol  {52.1} & \ccol {76.4} &\ccol  {85.2} &\ccol {51.2} &  \ccol  {77.9}   & \ccol {86.2} & \ccol {429.0}    \\
    \ccol  ~~+DSL & \ccol A+V+T &\ccol  \textbf{56.3} \textcolor{grn}{(+4.2)} & \ccol \textbf{80.8} \textcolor{grn}{(+4.4)} &\ccol  \textbf{88.1} \textcolor{grn}{(+2.9)} &\ccol \textbf{57.4} \textcolor{grn}{(+6.2)} &  \ccol  \textbf{80.2} \textcolor{grn}{(+2.3)}   & \ccol {87.4} \textcolor{grn}{(+1.2)} & \ccol \textbf{450.2} \textcolor{grn}{(+21.2)}    \\
[-0.3ex]\bottomrule 
\end{tabularx}
\vspace{-1.5mm}
\caption{Text-to-video and video-to-text retrieval results on the MSR-VTT 9k split. 
The post-processing techniques such as DSL~\cite{camoe}, QB-Norm~\cite{qbnorm}, and SK norm are used for further performance boosting.
}
\label{tab:supp_postprocessing}
\vspace{-2mm}
\end{table*}

\subsection{More Implementation Details}
\label{supp_sec:detail}
\bs{The details of the training configurations of our method across datasets are provided in Table~\ref{tab:config}.
We follow~\cite{clip4clip,eclipse} for most configurations, such as the image encoder, training epochs, optimizer, batch size, max frames, max words, learning rate for CLIP encoders, and temperature $\tau$. 
}

\subsection{More Quantitative Results}
\label{supp_sec:quan}
\noindent\textbf{Effect of Post-Processing:}
Post-processing techniques have been widely adopted in video-text retrieval to enhance performance by refining similarity scores.
Previous methods~\cite{camoe,ts2,ucofia,uatvr,tefal} adopt the post-processing techniques, including Dual Softmax Loss (DSL)~\cite{camoe}, Querybank Norm (QB-Norm)~\cite{qbnorm}, and the Sinkhorn-Knopp algorithm (SK-Norm), for further improvements in retrieval accuracy.
We also explore the effect of the post-processing technique by adopting DSL that applies inverted softmax~\cite{inverted} during inference.
We report the retrieval performance of AVIGATE with and without post-processing in Table~\ref{tab:supp_postprocessing} compared with existing methods.
Our model, AVIGATE, consistently achieves superior performance across all evaluation metrics for both text-to-video and video-to-text retrieval tasks, outperforming all previous methods by significant margins. 
Specifically, for the CLIP ViT-B/32 backbone, AVIGATE with post-processing achieves R@1 of 53.9\% for text-to-video retrieval.
Furthermore, in video-to-text retrieval, AVIGATE with DSL achieves R@1 of 53.0\%. 
Similarly, for the CLIP ViT-B/16 backbone, AVIGATE achieves substantial gains over existing methods. 
When using post-processing, our method achieves R@1 of 56.3\% for text-to-video retrieval, representing a considerable 2.3\%p improvement over TS2-Net~\cite{ts2}. 
In video-to-text retrieval, AVIGATE also outperforms other methods with R@1 of 57.4\%.

\noindent\textbf{Entire Performance on VATEX~\cite{vatex} and Charades~\cite{charades}:}
We present the complete video-text retrieval results on VATEX and Charades in Table~\ref{tab:supp_vatex}, including both text-to-video and video-to-text retrieval.
The results are reported using two variants of the CLIP ViT backbone, CLIP ViT-B/32 and CLIP ViT-B/16.
Moreover, we assess the effect of the post-processing technique, DSL~\cite{camoe}, for further performance boosts.
%
\bs{On VATEX, with the CLIP ViT-B/32 backbone, AVIGATE achieves notable results in text-to-video retrieval, with R@1 of 63.1\% and 76.6\% for text-to-video retrieval and video-to-text retrieval, respectively.
When applying DSL, we observe significant improvements across all metrics.
Specifically, it improves AVIGATE by a large margin, 7.6\%p and 8.7\%p in R@1 for text-to-video retrieval and video-to-text retrieval, respectively.
When using the larger backbone, CLIP ViT-B/16 backbone, AVIGATE demonstrates the scalability across different backbone sizes, achieving R@1 of 67.5\% for text-to-video retrieval and 80.7\% for video-to-text retrieval.
Moreover, the use of DSL consistently boosts the retrieval accuracy overall, with 20.2\%p improvements in RSum.
On Charades, with the CLIP ViT-B/32 backbone, AVIGATE achieves R@1 of 18.8\% in text-to-video retrieval and 17.2\% in video-to-text retrieval, which modestly increase to 21.3\% and 20.0\% when DSL is applied. 
Employing the larger backbone, CLIP ViT-B/16, AVIGATE attains R@1 of 24.1\% in text-to-video retrieval and 22.9\% in video-to-text retrieval, with DSL boosting these figures to 27.5\% and 27.1\%. 
}
\begin{table*}[!t]
\fontsize{8}{9.2}\selectfont
\centering
\begin{tabularx}{0.98\textwidth}
    {
      p{0.153\textwidth}
      >{\raggedright\arraybackslash}p{0.07\textwidth}
      >{\raggedright\arraybackslash}p{0.075\textwidth}
      >{\raggedright\arraybackslash}p{0.075\textwidth}
      >{\raggedright\arraybackslash}p{0.075\textwidth}
      >{\raggedright\arraybackslash}p{0.075\textwidth}
      >{\raggedright\arraybackslash}p{0.075\textwidth}
      >{\raggedright\arraybackslash}p{0.075\textwidth}
      >{\raggedright\arraybackslash}X
      }
     \toprule
    {\multirow{1}{*}[-4.5mm]{\textbf{Methods}}}& \multirow{1}{*}[-4.5mm]{\textbf{Modality}}&
    \multicolumn{3}{c}{\textbf{Text-to-Video Retrieval}} & \multicolumn{3}{c}{\textbf{Video-to-Text Retrieval}} & \multicolumn{1}{c}{\multirow{2}{*}{RSum}} \\ [-0.3ex]  \cmidrule(lr){3-5}  
    \cmidrule(lr){6-8} & & R@1 & R@5 & R@10 &  R@1 & R@5 & R@10 & \\  [-0.4ex] \midrule
    \multicolumn{9}{l}{\textit{\textbf{CLIP ViT-B/32}}} \\ [-0.3ex]\midrule
      AVIGATE (Ours) &  A+V+T &  {63.1} &  {90.7}  &  95.5 & 76.6 &    97.3   &  98.8 &  522.0    \\
      ~~+DSL &  A+V+T &  \textbf{70.7} \textcolor{grn}{(+7.6)} &  \textbf{93.4} \textcolor{grn}{(+2.7)} &  \textbf{95.5} \textcolor{grn}{(+1.4)} &  \textbf{85.3} \textcolor{grn}{(+8.7)} &  \textbf{99.1} \textcolor{grn}{(+1.8)} &  \textbf{99.8} \textcolor{grn}{(+1.0)} &  \textbf{545.2} \textcolor{grn}{(+23.2)} \\
    \midrule
    \multicolumn{9}{l}{\textit{\textbf{CLIP ViT-B/16}}} \\ [-0.4ex] \midrule

    AVIGATE (Ours) &  A+V+T &  67.5 &  93.2 &  96.7 & 80.7 &    97.8   &  99.5 &  535.4    \\
      ~~+DSL &  A+V+T &  \textbf{74.6} \textcolor{grn}{(+7.1)} &  \textbf{95.3} \textcolor{grn}{(+2.1)} &  \textbf{97.8} \textcolor{grn}{(+1.1)} &  \textbf{88.7} \textcolor{grn}{(+8.0)} &  \textbf{99.3} \textcolor{grn}{(+1.5)} &  \textbf{99.9} \textcolor{grn}{(+0.3)} &  \textbf{555.6} \textcolor{grn}{(+20.2)}          \\
\bottomrule 
\end{tabularx}
\vspace{-1.5mm}
\caption{Text-to-video and video-to-text retrieval results on VATEX. 
The post-processing technique, DSL~\cite{camoe}, is used for further performance boosting.
}
\label{tab:supp_vatex}
\vspace{-2mm}
\end{table*}

\begin{table*}[!t]
\fontsize{8}{9.2}\selectfont
\centering
\begin{tabularx}{0.98\textwidth}
    {
      p{0.153\textwidth}
      >{\raggedright\arraybackslash}p{0.07\textwidth}
      >{\raggedright\arraybackslash}p{0.075\textwidth}
      >{\raggedright\arraybackslash}p{0.075\textwidth}
      >{\raggedright\arraybackslash}p{0.075\textwidth}
      >{\raggedright\arraybackslash}p{0.075\textwidth}
      >{\raggedright\arraybackslash}p{0.075\textwidth}
      >{\raggedright\arraybackslash}p{0.075\textwidth}
      >{\raggedright\arraybackslash}X
      }
     \toprule
    {\multirow{1}{*}[-4.5mm]{\textbf{Methods}}}& \multirow{1}{*}[-4.5mm]{\textbf{Modality}}&
    \multicolumn{3}{c}{\textbf{Text-to-Video Retrieval}} & \multicolumn{3}{c}{\textbf{Video-to-Text Retrieval}} & \multicolumn{1}{c}{\multirow{2}{*}{RSum}} \\ [-0.3ex]  \cmidrule(lr){3-5}  
    \cmidrule(lr){6-8} & & R@1 & R@5 & R@10 &  R@1 & R@5 & R@10 & \\  [-0.4ex] \midrule
    \multicolumn{9}{l}{\textit{\textbf{CLIP ViT-B/32}}} \\ [-0.3ex]\midrule
      AVIGATE (Ours) &  A+V+T &  18.8 &  40.0  & 51.8  & 17.2 &   40.4  & 51.7  &  219.9 \\
      ~~+DSL &  A+V+T &  \textbf{21.3} \textcolor{grn}{(+2.5)} &  \textbf{42.4} \textcolor{grn}{(+2.4)} &  \textbf{54.4} \textcolor{grn}{(+2.7)} &  \textbf{20.0} \textcolor{grn}{(+2.8)} &  \textbf{43.0} \textcolor{grn}{(+2.6)} &  \textbf{54.9} \textcolor{grn}{(+3.2)} &  \textbf{236.0} \textcolor{grn}{(+16.1)} \\
    \midrule
    
    \multicolumn{9}{l}{\textit{\textbf{CLIP ViT-B/16}}} \\ [-0.4ex] \midrule

    AVIGATE (Ours) &  A+V+T &  24.1 &  48.5 &  61.3 & 22.9 &    48.4   &  61.0 & 266.2    \\
      ~~+DSL &  A+V+T &  \textbf{27.5} \textcolor{grn}{(+3.4)} &  \textbf{52.7} \textcolor{grn}{(+4.2)} &  \textbf{64.5} \textcolor{grn}{(+3.2)} &  \textbf{27.1} \textcolor{grn}{(+4.2)} &  \textbf{52.7} \textcolor{grn}{(+4.3)} &  \textbf{65.0} \textcolor{grn}{(+4.0)} &  \textbf{289.5} \textcolor{grn}{(+23.3)}          \\
\bottomrule 
\end{tabularx}
\vspace{-1.5mm}
\caption{\bs{Text-to-video and video-to-text retrieval results on Charades. 
The post-processing technique, DSL~\cite{camoe}, is used for further performance boosting.}
}
\label{tab:supp_char}
\vspace{-2mm}
\end{table*}


\subsection{More Ablation Studies}
\label{supp_sec:ablation}
We further conduct ablation studies using varying hyperparameters in AVIGATE.
Similar to the main paper, we report text-to-video retrieval results on the MSR-VTT dataset~\cite{msrvtt} with CLIP ViT-B/32. Table~\ref{tab:supp_ablation_hyp} presents the whole results of the ablation studies.

\noindent \textbf{Layer Depth of Gated Fusion Transformer:}
We present the impact of the number of layers of the gated fusion transformer ($L$) in Table~\ref{tab:supp_ablation_hyp}(a)  and observe that the performance gradually improves up to $L$=4, where the best performance is achieved.

\noindent \textbf{Hyperparameters $\lambda$ and $\delta$ in Eq.~(5):}
We investigate the impact of the scaling factor $\lambda$ and the maximum margin $\delta$ in Eq.~(5) of the manuscript.
It is worth noting that the adaptive margin in Eq.~(5) becomes 0 when $\lambda$ or $\delta$ are set to 0, leading the loss in Eq.~(6) to the conventional contrastive loss.
As shown in Table~\ref{tab:supp_ablation_hyp}(b), when $\lambda$ is set to 0.2, the model yields the best performance. 
Meanwhile, setting $\lambda$ to 0.1 results in a slight decrease in performance, indicating that a smaller scaling factor may not provide sufficient margin adjustment.  
However, increasing $\lambda$ to 0.3 does not lead to further improvements.
Similarly, Table~\ref{tab:supp_ablation_hyp}(c) presents the effect of varying the maximum margin $\delta$. 
We observe that the performance gradually improves up to $\delta = 0.1$.
Increasing $\delta$ beyond 0.1 degrades performance due to excessively large margins pushing negative pairs too far apart.

\begin{table}[!t]
\setlength{\tabcolsep}{2pt}
\fontsize{8.2}{9}\selectfont
\centering
\scalebox{0.95}{
\begin{tabularx}{0.499\textwidth}
    {
      >{\centering\arraybackslash}p{0.13\textwidth}
      >{\centering\arraybackslash}X
      >{\centering\arraybackslash}X
      >{\centering\arraybackslash}X
      }
     \toprule
    \multirow{1}{*}[-4.5mm]{\textbf{Ablated Setting}} & \multicolumn{3}{c}{\textbf{Text-to-Video Retrieval}} \\ [-0.3ex] \cmidrule(lr){2-4}  
    &  R@1 & R@5 & R@10   \\  [-0.4ex] \midrule
    \multicolumn{4}{l}{\textit{\textbf{(a) Layer depth of Gated Fusion Transformer: $L$}}} \\ [-0.3ex]\midrule
    $L$=1 & 49.0 & 74.0 & 82.6 \\
    $L$=2  & 49.8 & 74.0 & 83.0 \\
    \ccol$L$=4  &\ccol 50.2 &\ccol  74.3 & \ccol 83.2 \\
    $L$=6  & 49.5 & 74.2 & 82.6 \\    [-0.3ex]\midrule
    \multicolumn{4}{l}{\textit{\textbf{(b) Scaling factor in Eq.(5): $\lambda$}}} \\ [-0.3ex]\midrule
    $\lambda$=0.0 & 48.0 & 75.1 & 83.4 \\
    $\lambda$=0.1 & 49.4 & 74.8 & 83.8 \\
    \ccol$\lambda$=0.2  &\ccol 50.2 &\ccol  74.3 & \ccol 83.2 \\
    $\lambda$=0.3  & 50.0 & 74.4 & 83.2 \\    [-0.3ex]\midrule
    \multicolumn{4}{l}{\textit{\textbf{(c) Maximum margin in Eq.(5): $\delta$}}} \\ [-0.3ex]\midrule
    $\delta$=0.00 & 48.0 & 75.1 & 83.4 \\
    $\delta$=0.05 & 49.4 & 75.1 & 83.6 \\
    \ccol$\delta$=0.10  &\ccol 50.2 &\ccol  74.3 & \ccol 83.2 \\
    $\delta$=0.15 & 49.3 & 74.8 & 83.8 \\
    $\delta$=0.20  & 48.3 & 74.4 & 83.9 \\  [-0.3ex]\midrule
    \multicolumn{4}{l}{\textit{\textbf{(d) Gate mechanism type}}} \\ [-0.3ex]\midrule
    Hard Gate & 49.3 & 75.0 & 82.5 \\
    \ccol Soft Gate  &\ccol 50.2 &\ccol  74.3 & \ccol 83.2 \\ [-0.3ex]\midrule 
    \multicolumn{4}{l}{\textit{\textbf{(e) Effect of freezing AST (Batch size:32)}}} \\ [-0.3ex]\midrule
    Freezing  &  48.2 & 75.3 &  83.7  \\
    Fine-tuning &  48.0 & 73.5 & 83.4  \\ \midrule
    \multicolumn{4}{l}{\textit{\textbf{(f) Effect of freezing CLIP encoders }}} \\ [-0.3ex]\midrule
    Freezing  & 41.1 & 68.5  & 78.2  \\ 
    \ccol Fine-tuning & \ccol 50.2 & \ccol 74.3 & \ccol 83.2  \\\midrule    
[-0.3ex]\bottomrule 
\end{tabularx}
}
\vspace{-1.5mm}
\caption{Ablation studies on hyperparameters. 
\colorbox{grey}{{gray}} corresponds to our default setting.
}
\label{tab:supp_ablation_hyp}
\vspace{-2mm}
\end{table}

\noindent \textbf{Gate Mechanism Type:}
Our method employs a soft gate mechanism, which allows for continuous modulation of the contribution of audio during fusion.
To evaluate the effectiveness of the soft gate mechanism, we compare the soft gate with a hard gate mechanism, which assigns a gating score of 1 if it exceeds a predefined threshold and 0 otherwise.
As shown in Table~\ref{tab:supp_ablation_hyp}(d), using the hard gate underperforms our method.
Unlike using the hard gate mechanism, our method facilitates the effective use of relevant audio cues while minimizing the impact of irrelevant or noisy audio signals; it enables the model to leverage informative audio more precisely, thereby improving retrieval accuracy.

\bs{\noindent \textbf{Effect of freezing AST:} 
We freeze AST to reduce training costs.
Fine-tuning AST is impractical since it processes 1,214 tokens per input audio, far more than 50 tokens for each video frame in ViT-B/32.
A solution is to largely reduce the batch size, which however degrades performance since the contrastive loss is highly dependent on the batch size.
The results of freezing and fine-tuning AST with tiny input batches are reported in Table~\ref{tab:supp_ablation_hyp}(e), while freezing AST outperforms fine-tuning it.
The results are attributed to the characteristics of AST pre-trained on the audio classification dataset, allowing it to extract discriminative embeddings from audio inputs.
Therefore, we decided to freeze the AST instead of fine-tuning that requires a burden of computational and memory costs.

\noindent \textbf{Freezing both CLIP image and text encoders:}
As shown in Table~\ref{tab:supp_ablation_hyp}(f), freezing the CLIP image and text encoders leads to noticeably lower performance, highlighting the importance of fine-tuning both encoders, as also demonstrated in prior work such as CLIP4Clip~\cite{clip4clip}.
Fine-tuning is essential for capturing task-specific video and text information and improving the alignment between them.}

\subsection{More Qualitative Results}
\label{supp_sec:qual}
\begin{figure*}[t!]
    \vspace{-1mm}
    \centering
    \includegraphics[width=0.99\linewidth]{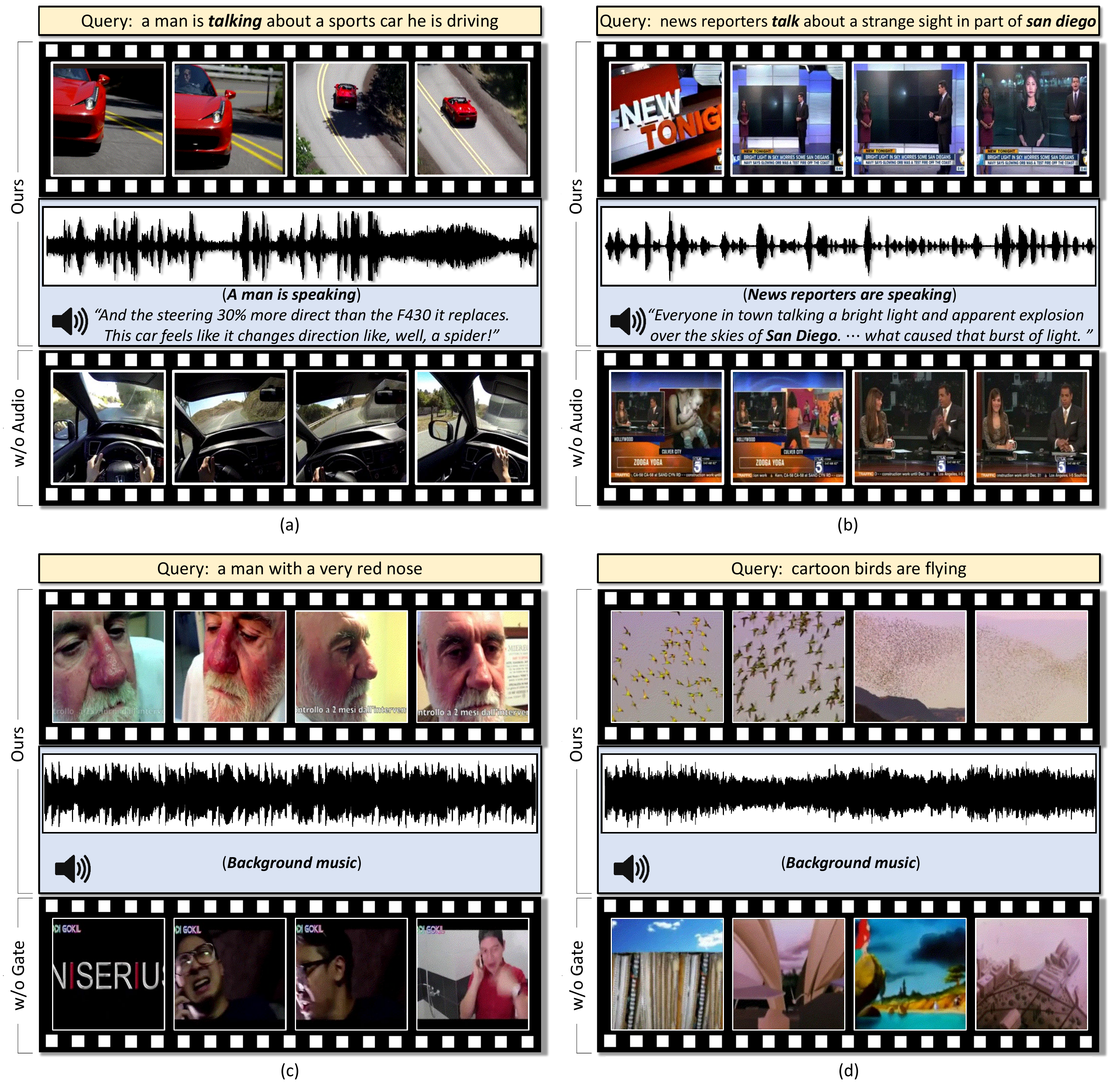}
    \vspace{-2mm}
    \caption{ Top-1 text-to-video retrieval results of our method on MSR-VTT, where they are true matches.
    The audio provides informative cues for accurate retrieval, where ``a man is talking'' in the query text is not visible (a) and ``talk$\cdots$san diego'' in the query text is not visible but audible (b).
    However, neglecting these informative audio signals (\ie, w/o Audio) fails to retrieve true matches. 
    Meanwhile, the irrelevant audio is filtered by the gated fusion transformer, leading to accurate retrieval results (c) and (d); without the gating mechanism (\ie, w/o Gate), it leads to retrieving false matches due to the irrelevant audio.    
    }
    \label{fig:supp_qual}
    \vspace{-3.5mm}
\end{figure*}

We further present additional qualitative results that illustrate the effectiveness of AVIGATE in leveraging audio information for text-to-video retrieval. 
Figure~\ref{fig:supp_qual} shows the Top-1 retrieved videos from our method, including the corresponding audio signals, to highlight how audio cues influence retrieval outcomes.

In Figure~\ref{fig:supp_qual}(a) and (b), we present a scenario where the audio provides valuable information that enables improving retrieval performance. 
AVIGATE, which incorporates audio through the gated fusion transformer, successfully retrieves the correct video corresponding to the text query. In contrast, the method without audio information (\ie, w/o Audio) fails to retrieve the true matches. This comparison highlights the benefit of utilizing informative audio cues.

Conversely, Figure~\ref{fig:supp_qual}(c) and (d) present another scenario where the audio input contains irrelevant information, such as background noise. AVIGATE effectively filters out the uninformative audio signals through the gating mechanism. The gating function assigns low gating scores, allowing the model to focus only on the visual cues. As a result, AVIGATE successfully retrieves the correct videos. In contrast, the method without the gating function (\ie, w/o Gate) is impacted by the noisy audio and fails to retrieve the true matches.

These qualitative results demonstrate that the gated fusion transformer successfully filters out irrelevant audio while leveraging valuable audio information when the audio contributes positively.

\end{document}